\documentclass{article}

\usepackage{microtype}
\usepackage{graphicx}
\usepackage{subfigure}
\usepackage{booktabs} % for professional tables

\usepackage{hyperref}
\usepackage{colortbl}
\usepackage{xcolor}
\colorlet{shadecolor}{gray!20}

\usepackage[accepted]{icml2023}

\usepackage{amsmath}
\usepackage{amssymb}
\usepackage{mathtools}
\usepackage{amsthm}

\usepackage[capitalize,noabbrev]{cleveref}

\theoremstyle{plain}

\theoremstyle{definition}

\theoremstyle{remark}

\usepackage[textsize=tiny]{todonotes}

\usepackage{amsfonts}

\usepackage{graphicx}

\usepackage{color}
\usepackage{multirow}

\usepackage{bbm}
\usepackage{enumitem}
\usepackage[flushleft]{threeparttable}
\usepackage{makecell}
\renewcommand{\algorithmicrequire}{\textbf{Input:}}
\renewcommand{\algorithmicensure}{\textbf{Output:}}

\newcommand{\secref}[1]{\S~\ref{#1}}
\newcommand{\figref}[1]{Figure~\ref{#1}}
\newcommand{\eqnref}[1]{Eq.~(\ref{#1})}
\newcommand{\tabref}[1]{Table~\ref{#1}}
\newcommand{\tabincell}[2]{\begin{tabular}{@{}#1@{}}#2\end{tabular}}

\icmltitlerunning{In-Context Learning with Many Demonstration Examples}

\begin{document}

\twocolumn[
\icmltitle{In-Context Learning with Many Demonstration Examples}



\icmlsetsymbol{equal}{*}

\begin{icmlauthorlist}
\icmlauthor{Mukai Li}{shlab}
\icmlauthor{Shansan Gong}{shlab}
\icmlauthor{Jiangtao Feng}{shlab}
\icmlauthor{Yiheng Xu}{shlab,hku}\\
\icmlauthor{Jun Zhang}{shlab}
\icmlauthor{Zhiyong Wu}{shlab}
\icmlauthor{Lingpeng Kong}{shlab,hku}
\end{icmlauthorlist}

\icmlaffiliation{shlab}{Shanghai Artificial Intelligence Laboratory}
\icmlaffiliation{hku}{Department of Computer Science,The University of HongKong}

\icmlcorrespondingauthor{Jiangtao Feng}{fengjiangtao@pjlab.org.cn}
\icmlcorrespondingauthor{Lingpeng Kong}{lpk@cs.hku.hk}

\icmlkeywords{In-context Learning, Pre-training Language Model, Instruction Tuning}

\vskip 0.3in
]



\printAffiliationsAndNotice{}  % leave blank if no need to mention equal contribution

\begin{abstract}
Large pre-training language models (PLMs) have shown promising in-context learning abilities.
% which can substantially be further improved by instruction tuning. 
However, due to the backbone transformer architecture, existing PLMs are bottlenecked by the memory and computational cost when scaling up to a large context size, leaving instruction tuning and in-context learning of many demonstration examples, as well as long-range language modeling under-explored.
% Due to the lack of PLM trained on long contexts, along with the poor extrapolation ability of current PLMs, instruct tuning and in-context learning with more demonstration examples have remained to be studied. 
In this study, we propose a long-range language model \textsc{EvaLM} based on an efficient transformer mechanism. \textsc{EvaLM} is trained with $8$k tokens per batch line and can test up to $256$k-lengthed contexts with extrapolation,  $128\times$ to the limit of existing PLMs (e.g. GPT3).
Based on \textsc{EvaLM}, we scale up the size of examples efficiently in both instruction tuning and in-context learning to explore the boundary of the benefits from more annotated data.
% make better use of annotated data. 
Experimental results on a diverse set of tasks show that \textsc{EvaLM} achieves 4.1\% higher accuracy on average, and the average length of achieving the best accuracy score over tasks is around 12k. We find that in-context learning can achieve higher performance with more demonstrations under many-shot instruction tuning (8k), and further extending the length of instructions (16k) can further improve the upper bound of scaling in-context learning. Code is available on \url{https://github.com/Shark-NLP/EVALM}.

\end{abstract}
\section{Introduction}
% Large pre-training language models have 

% But Long Range Language Model~(LRLM) as well as its in-context learning capability remains unexplored.
%brief outline
%1 PLM gain emergent ability with the increasing model size，icl learning have become popular。especially with the help of instructing learning the icl ability become even stronger
% but due to the length strict，scaling up the in context length have not be well studied。 most the existing model trained on max length 2048， the demostration number have upper bond。also limit the icl num
% thus we design evalm，…………………………………………。
%

 With the increasing scale of pre-trained language models (PLMs), in-context learning (ICL) has emerged as a novel paradigm for utilizing PLMs~\citep{brown2020language,zhang2022opt,chowdhery2022palm}. Unlike learning methods that require updating parameters, in-context learning allows for good model performance with a prompt that only includes natural language instructions and/or a few demonstrations~\citep{dong_survey}. 
 % Recent research starts to explore the effects of sample selection, sample quantity, and order on PLM performance in in-context settings. Additionally, 
 In addition to that, a recent line of research on instruction tuning shed new light on closing the gap between pre-training and in-context learning~\citep{chung_scaling_2022, min-etal-2022-metaicl}, facilitating the usage of natural language instructions to interact with the PLMs.

However, the computational overhead of the backbone vanilla transformer architecture prevents existing PLMs from a longer context. A maximum context size (i.e., 2048) is set in the most popular pre-training models (e.g., GPT3, \citealt{brown2020language}; OPT, \citealt{zhang2022opt}; PaLM \citealt{chowdhery2022palm}). 
The direct consequence is scaling up to large numbers of samples in instruction tuning or in-context learning becomes under-explored. How effectively can we improve the in-context learning performance of the PLMs by serving more demonstration examples?
% The limitation of context size also prevent instruction tuning and in-context learning from scaling to more samples, making the exploration of scaling up in-context learning a remained problem.

To answer this question, we start from responding to the challenge of long-range language models (LRLMs). We train an LRLM named \textsc{EvaLM} (\secref{sec:learning}), which backbones on a state-of-the-art efficient transformer architecture \textsc{EVA}~\citep{zheng2023efficient}, with modifications to handle the extrapolation of position embeddings (\secref{sec:arch}). \textsc{EvaLM} with many-shot instruction tuning achieves better performance in long-range language modeling with cheap memory and computational costs (\secref{sec:discuss}).
% typical Transformer~\citep{vaswani2017attention} decoder, with the modification of attention mechanism for efficiency and position embedding for extrapolation. 
% Specifically, we adopt a causal version of EVA~\citep{anonymous2023efficient}, which is an efficient attention mechanism that achieves comparable performance to vanilla attention but uses less memory and computational costs. 
% Besides, EVA gains its ability from efficiency to tackle with longer context than its peers. 
The learned circular position embedding and incremental encoding we propose help \textsc{EvaLM} to extrapolate to an input length of $256$k tokens effectively. 
We then conduct a series of experiments testing the performance of \textsc{EvaLM} when scaling up the number of demonstration examples in ICL in various tasks. We find that with more demonstration examples, \textsc{EvaLM} is able to achieve better ICL performance than comparable PLMs with rare extra overheads.
% can improve model performance in various tasks. Additionally, using more samples for instruction tuning can also effectively improve in-context performance. Using these strategies and the above techniques based on our \textsc{EvaLM}, we are able to achieve better ICL performance than comparable PLMs with rare extra overheads. 
We summarize our contribution as follows:
\begin{enumerate}
\setlength\itemsep{-0.4em}

\item We pre-train a long-range language model named \textsc{EvaLM} using similar training costs with OPT, which enables the scale-up of instruction tuning and in-context learning, showing that many-shot instruction tuning can help ICL achieve higher performance with larger demonstrations, and with longer instructions, this phenomenon is more obvious. However, this increasing trend is not endless.
\item We enable many-shot instruction tuning and in-context learning inference with the cooperation of incremental encoding to ensure efficiency and circular position embedding to ensure extrapolation.
\item We conduct experiments on 10 commonly used datasets covering diverse tasks with different prompting strategies. \textsc{EvaLM} with many-shot instruction tuning achieves 4.1\% higher accuracy on average, at 12k inputs length on average.

\end{enumerate}

\section{Related Work}
\paragraph{Pre-trained Language Model}
PLMs are trained on large and general corpora and then finetuned or few-shot transferred to perform various NLU and NLG tasks. Among them, besides encoder-decoder Transformer~\citep{vaswani2017attention} architecture such as T5~\citep{raffel2020exploring}, there are auto-regressively pre-trained models, like XLNet~\citep{yang2019xlnet}, GPT~\citep{radford2019language, brown2020gpt3, black_gpt-neox-20b_2022}, OPT~\citep{zhang2022opt}, PaLM~\citep{chowdhery2022palm}, BLOOM~\citep{scao-2022-bloom}, and etc. These decoder-based causal language models soon occupy kinds of NLP leaderboards, showing excellent language modeling and in-context learning ability of them. However, the huge computing overhead (including memory and time consumption) makes nonprofits and smaller labs difficult to create or even use PLMs. Furthermore, this also prevents PLMs from encoding longer inputs. 
% H3~\cite{dao2022hungry} improves the efficiency of training by ameliorating the state space model to replace the original attention and is able to scale up to the sequence length of 8K.
% Evidence in GPT3~\citep{brown2020gpt3} shows that these large-scale models can learn in-context through few-shot learning. 

\paragraph{Efficient Attention}
A surge of efficient attention models are devised to enhance the efficiency of the original Transformer model~\citep{vaswani2017attention}.
These models explore diverse philosophies to improve the efficiency, including sparse attention matrix~\citep{luong2015effective, tay2020sparse, beltagy2020Longformer, zaheer2020big, ainslie2020etc}, memory compression~\citep{liu2018generating, lee2019set, rae2020Compressive, wang2020linformer} low-rank decomposition~\citep{xiong2021nystromformer, lu2021soft, chen2021skyformer}, kernel-based linear attention~\citep{choromanski2021rethinking, peng2021random, peng2022abc, zheng2022linear, zheng2023efficient}, state-space model~\citep{gu2022efficiently, gupta2022diagonal, dao2022hungry}, and CUDA re-implementation~\citep{dao2022flashattention}. Thus models with efficient attention architecture are promising to handle longer input sequences when memory consumption is saved. H3~\cite{dao2022hungry} is pre-trained as an efficient language model but fails to scale up the training sequence length which remains 2048.

\paragraph{In-Context Learning}
With the increasing scale and capacity of PLMs, ICL has become a new paradigm for NLP~\citep{brown2020gpt3}. The success of ICL has been demonstrated on a wide range of NLP tasks, including question answering~\cite{joshi2017triviaqa}, information retrieval~\cite{tay2022transformer}, math word problem~\cite{cobbe2021training}, commonsense reasoning~\cite{geva2021did}, and fact checking~\cite{rae2021scaling} etc. Several recent studies~\cite{liu2022makes,wu2022self} have observed a positive correlation between the number of in-context examples and ICL's performance: increasing the number of in-context examples can bring steady improvements. 
Further investigation is carried out to pack and/or distill more examples into the context through continued pre-training~\cite{choi2022prompt}, and instruction tuning~\cite{snell2022learning}. However, the input length limitation of current PLMs still restricts us from directly feeding more in-context examples into the model.

\section{\textsc{EvaLM}}

\begin{figure*}[t]
 \centering
 \includegraphics[width=1.0\textwidth]{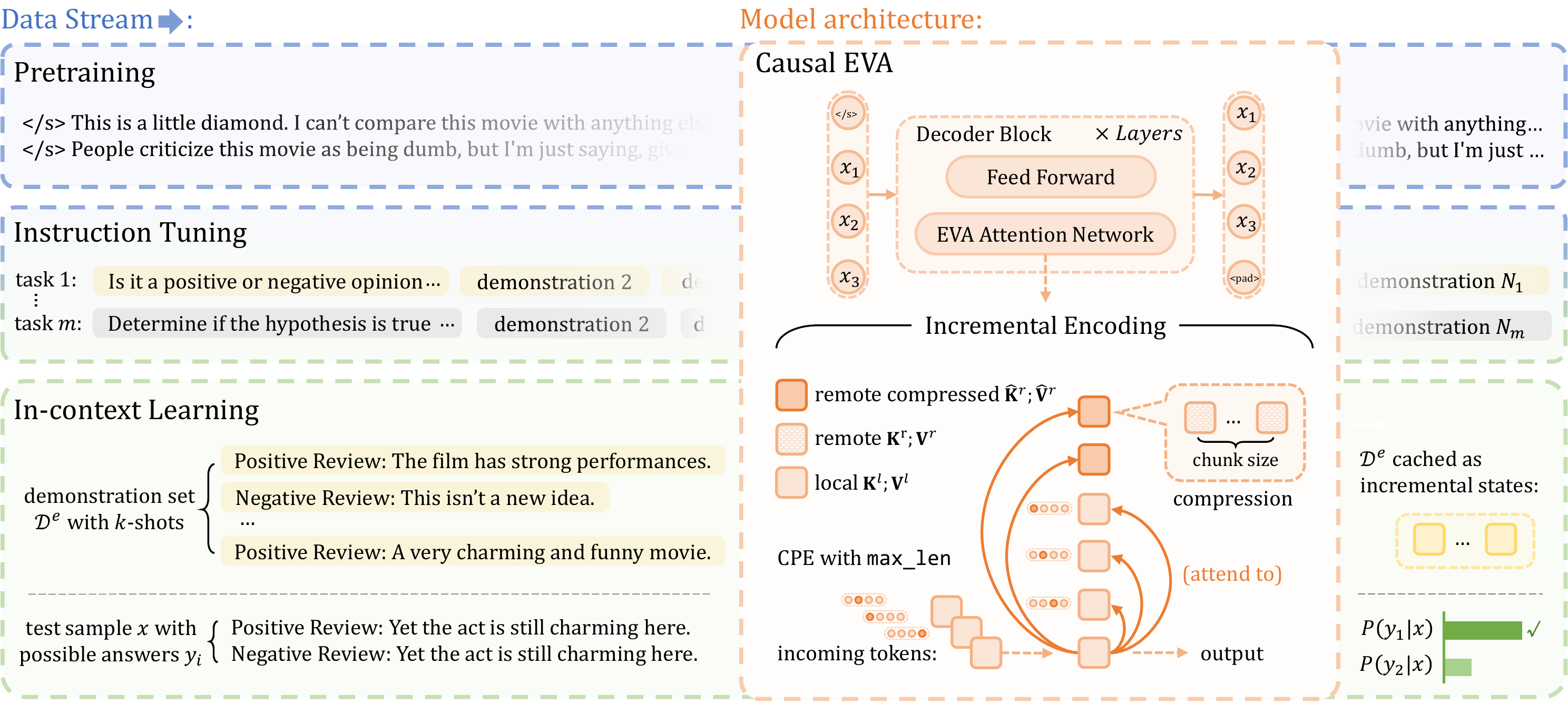}
 \caption{The illustration of \textsc{EvaLM} scaling up in-context learning. The pre-training stage empowers the language modeling capacity of \textsc{EvaLM}, and the instruction tuning explicitly aligns \textsc{EvaLM} with instructions from different tasks. For different downstream tasks, \textsc{EvaLM} can in-context learn from the demonstrations. With the help of CPE and incremental encoding technique, $k$ could be scaled up.}
 \label{fig:evalm}
\end{figure*}

We propose a long-range language model named \textsc{EvaLM} to scale up the sequence length reached by existing pre-trained language models.
The rest of this section is organized as follows: 
\S~\ref{sec:arch} introduces the overall architecture of \textsc{EvaLM}; 
\S~\ref{sec:learning} focuses on learning \textsc{EvaLM} on both of pre-training and instruction tuning;
\S~\ref{sec:icl} shows how \textsc{EvaLM} scales up the maximum size of shots in in-context learning, with an incremental encoding technique. The overall architecture is shown in~\figref{fig:evalm}.

\subsection{Architecture}
\label{sec:arch}
We adopt EVA~\citep{zheng2023efficient}, a recently introduced attention competitor, as an efficient alternative to vanilla softmax attention~\citep{vaswani2017attention}, for its high efficiency in long sequence modeling and strong performance.
The original EVA performs both causal and noncausal attention in sequence modeling, and here we focus on its causal version for its adaption to language modeling.
A general computation process of causal EVA is described as follows. 
Given a query $\mathbf{q}_t \in \mathbb{R}^d$, and key-value sequences $\mathbf{K}_{1:t}, \mathbf{V}_{1:t} \in \mathbb{R}^{t \times d}$, where $d$ is the dimensionality and $t$ is the timestamp, EVA learns attentive features as:
% \begin{enumerate}
% \setlength\itemsep{-0.3em}
  a) chunking key-value features $\mathbf{K}_{1:t}, \mathbf{V}_{1:t}$ as $\mathbf{K}^r, \mathbf{K}^l=C(\mathbf{K}_{1:t}),\mathbf{V}^r, \mathbf{V}^l=C(\mathbf{V}_{1:t})$, where $C(\cdot)$ is a chunking function with chunk size $c$, and superscripts $r$ and $l$ denote the remote features beyond present chunk of $\mathbf{q}_t$ and the local features within the chunk;
  b) compressing remote features within each chunk by another efficient attention and pooling operation $\mathcal{M}(\cdot)$ as $\hat{\mathbf{K}}^r=\mathcal{M}(\mathbf{K}^r), \hat{\mathbf{V}}^r=\mathcal{M}(\mathbf{V}^r)$, where the efficient attention here is LARA~\citep{zheng2022linear};
  c) performing vanilla attention on concatenated remote and local features by 
  % \begin{equation}
  $\text{EVA}(\mathbf{q}_t)=\mathrm{softmax}(\mathbf{q}_t [\hat{\mathbf{K}}^r;\mathbf{K}^l]^\top)[\hat{\mathbf{V}}^r;\mathbf{V}^l]^\top$.
  % \end{equation}
% \end{enumerate}
% where the superscript $g$ and $l$ denotes global and local features respectively.
% The $\mathbf{K}^g$ and $\mathbf{V}^g$ are compressed from the chunks beyond the current chunk of $\mathbf{q}_t$ by a another efficient attention as ; the $\mathbf{K}^l$ and $\mathbf{V}^l$ is directed copied from the $\mathbf{q}_t$'s chunk as blockwise local attention~\citep{alrfou2019character, dai2019transformer}.  
% EVA is designed based on chunking methods, which splits sequences into multiple chunks with the same length.
% The key idea of EVA is jointly normalizing the global and local features, where the global features are compressed from remote chunks with another efficient attention as RFA~\citep{peng2021random} or LARA~\citep{zheng2022linear}, and the local features are borrowed from features within the same chunk.  
It is worth noting that EVA is capable of handling long-term dependencies by performing attention on remote compressed features $\hat{\mathbf{K}}^r, \hat{\mathbf{V}}^r$.
We refer interested readers to \citep{anonymous2023efficient} for further details.

Apart from the advanced attention mechanism EVA, we present circular positional embedding~(CPE) to enforce position information.
For $i$-th token, its positional embedding is set to $\mathbf{p}_{i\%M} \in \mathbb{R}^d$, where $M$ is the maximum size of learned positional embeddings.
An intriguing characteristic of CPE is its ability on extrapolation and long-term dependency.
CPE implicitly learns a position-aligned matrix $\mathbf{P}=\{\mathbf{p}_{i\%M}^\top \mathbf{p}_{j\%M}\}$ between each pair of tokens, which is added to attention matrices.
The matrix $\mathbf{P}$ is close to the pattern of strided attention~\citep{ho2019axial, tay2020sparse} with stride size $M$, and encourages feature interaction to distant features.

\paragraph{Extrapolation}
Extrapolation is a vital challenge in long-range language modeling.
Remind that the LRLMs are expected to scale the sequence length to tens, hundreds, or even more times to the current limitation with thousands of tokens from existing mainstream pre-trained language models such as GPT~\citep{brown2020gpt3} and OPT~\citep{zhang_opt_2022}.
The challenges of LRLMs lie in the two aspects.
On the one hand, such length is still unaffordable for current models, even for efficient attention models, during the training stage, despite incremental decoding~\citep{ott2019fairseq} helping reduce memory consumption in the inference stage.
On the other hand, pre-trained data from long-range texts are limited. 
Thus a practical solution is ``train short, test long'', a.k.a. extrapolation.
Thus finding an architecture with extrapolation capability is important for LRLMs.
In \textsc{EvaLM}, we enhance its extrapolation in two aspects:
a) based on the observation that locality contributes to extrapolation~\citep{zhang2022cab}, we choose EVA that also models the locality;
b) we use circular positional embedding that fledges vanilla learned positional embedding to extrapolate to longer contexts.
% we use sinusoidal position embedding~\citep{vaswani2017attention} instead of widely-used learnt one~\citep{brown2020language, chowdhery2022palm, zhang2022opt}, although learnt one obtains impressive performance on downstream tasks.

\subsection{Pre-training \& Instruction Tuning}
\label{sec:learning}
We pre-train a causal language model \textbf{\textsc{EvaLM}} based on EVA transformer decoder with our preprocessed Pile~\citep{Gao2020ThePA} corpus and further tune it using Many-Shot Instruction Tuning (MSIT).

\paragraph{Pre-training} Data processing details are in Appendix~\ref{appendix:data}, Pre-training details are in Appendix~\ref{appendix:training}. Our \textsc{EvaLM} was trained on a widely-used corpus the Pile~\citep{Gao2020ThePA}, which is a massive dataset designed for training large language models.
We built a preprocessing pipeline including filtering, deduplicating, and blending to prepare the pre-training corpus to support the large-scale distributed training process. We conducted catalog and content filtering following BLOOM~\citep{scao-2022-bloom} and deduplicated the filtered data using fuzzy deduplication similar to previous work~\citep{zhang_opt_2022, Smith2022UsingDA}.
The final corpus roughly contains 121B tokens. Please refer to Appendix~\ref{appendix:data} for detailed data processing and comparison. 

The training process \textsc{EvaLM} mainly follows GPT3~\citep{brown2020gpt3} and OPT~\citep{zhang2022opt}, optimizing the negative log-likelihood of next tokens in an auto-regressive way. We scaled the training sequence length to 8192 to accommodate more in-context samples. Fully sharded data parallel (FSDP) was applied in our pre-training stage, which can reduce the memory footprint of a single GPU to accommodate longer sequences. 
Please refer to Appendix~\ref{appendix:training} for a detailed pre-training setting. 
\paragraph{Many-Shot Instruction Tuning}
Instruction tuning simulates the in-context learning settings and shows the promising ability to activate the model's respective capacity during inference~\citep{min-etal-2022-metaicl}, with maximum training shots to 32. Based on our long-range \textsc{EvaLM}, we can further investigate the impact of instruction tuning after scaling up the shots. 
We instruction-tuned \textsc{EvaLM} on $m$ instruction tuning tasks $\mathcal{D}^{IT}_j=\{(\boldsymbol{x}_i^j,\boldsymbol{y}_i^j)\}_{i=1}^{N_j}$, which $m$ is the number of tasks and $N_j$ is sample number for each dataset.
Each input-output pair $(\boldsymbol{x}_i^j,\boldsymbol{y}_i^j)$ is turned into an instruction sequence $\boldsymbol{s}^{IT}_i=\mathcal{I}(\boldsymbol{x}_i, \boldsymbol{y}_i)$ wrapped by the instruction $\mathcal{I}(\cdot)$ written in natural language. $\mathcal{I}(\cdot)$ is derived from an instruction templates pool that is manually designed for different tasks $j$. 
Before feeding into \textsc{EvaLM}, we concatenate instructions until their total length reaches the limitation of 8192, in a batch-by-token way, named many-shot instruction tuning (MSIT). Further, we introduce the plus version of MSIT, which extrapolates the total length of instructions per batch line to $2\times 8192$.
% For each iteration, we obtain training batches from $\mathcal{T}_i:(x_1,y_1),...,(x_K,y_K)$ and wrap them with several natural text of instructions. 
The pressed \textsc{EvaLM} learns the concatenated $\boldsymbol{s}^{IT}_i$, under the supervision of the negative log-likelihood objective as language modeling. The data used in our instruction tuning refers to Appendix~\ref{appendix:IT}.
% \fjt{stop here}
% Different from previous work~\citep{min-etal-2022-metaicl} which fixed instruction shot $K$ the same with the number of in-context learning $k$, we propose 3 strategies to choose $K$: (1) \textbf{Shot-first}: using fixed but larger $K$; (2) \textbf{Shot-mixed}: mixed different $K$; (3) \textbf{Length-first}: increase $K$ until reaching maximum length.

\subsection{In-Context Learning with \textsc{EvaLM}}
\label{sec:icl}
% \paragraph{ICL Formulation}
% For a task-specific dataset, there is a set of ($k$) input-output examples $\{(\boldsymbol{x}_i, \boldsymbol{y}_i)|i\in [k]\}$ from the training set.
Consider a downstream task with dataset $\mathcal{D}$, from which we construct a demonstration exemplar set $\mathcal{D}^e=\{(\boldsymbol{x}_i, \boldsymbol{y}_i)\}_{i=1}^{k}$.
As instruction tuning, the demonstration exemplars are turned into instruction sequences and then concatenated to $\boldsymbol{s}^e=[\mathcal{I}(\boldsymbol{x}_i, \boldsymbol{y}_i)]_{i=1}^{k}$.
% Given a template of a task instruction $\mathcal{I}$,
% We construct a demonstration set $\mathcal{D}^d=\{d_1, d_2, ..., d_k\}$ where $d_i=\mathcal{I}(\boldsymbol{x}_i,\boldsymbol{y}_i)$ is a demonstration written in natural language text based on the template wrapper. 
% \fjt{stop here}
For a test input text $\boldsymbol{x}$ and its corresponding candidate categories $\mathcal{Y}$, we first concatenate $\boldsymbol{s}^e$ and $\mathcal{I}(\boldsymbol{x}, \boldsymbol{y})$ together to form a prompt for $\boldsymbol{y} \in \mathcal{Y}$. The prompt is then fed into the pre-trained \textsc{EvaLM} to compute the likelihood of the current answer $\boldsymbol{y}$ along with $\boldsymbol{x}$, and we choose the most possible one as the predicted label:

\begin{equation}
\label{eq:ppl}
  % P(Y_i|x)=\sum_{t\in l} \log p_t([D^*,\mathcal{I}(x, Y_i)]^l),
  % \arg \max_{y \in \mathcal{Y}} P(\mathcal{I}(x, y)|\boldsymbol{s}^e)
  \arg \max_{\boldsymbol{y} \in \mathcal{Y}} P([\boldsymbol{s}^e;\mathcal{I}(\boldsymbol{x}, \boldsymbol{y})]).
\end{equation}
There are several approaches to constructing $\mathcal{D}^e$ specifically, listed in~\secref{exp:setup}. For instance-level ICL, the same test sample $\boldsymbol{x}$ shares the same $\boldsymbol{s}^e$, and for dataset-level ICL, all test samples share the same $\boldsymbol{s}^e$~\citep{wu2022self}. In this situation, \eqnref{eq:ppl} turns into:

\begin{equation}
  \label{eq:ppl-incre}
  % \begin{split}
    % &\arg \max_{y \in \mathcal{Y}}\left( P(\boldsymbol{s}^e) + P(\mathcal{I}(x, y)|\boldsymbol{s}^e)\right)\\
   \arg \max_{\boldsymbol{y} \in \mathcal{Y}}P(\mathcal{I}(\boldsymbol{x}, \boldsymbol{y})|\boldsymbol{s}^e).
  % \end{split}
\end{equation}

% To construct $D$, following \citeauthor{liu-etal-2022-makes} and \citeauthor{gao-etal-2021-making}, we choose samples with top-$k$ similarity between the input $\boldsymbol{x}_i$ and the test sample $x$ in the embedding space, denoted as $s(\boldsymbol{x}_i,x)$. We follow $s(\boldsymbol{x}_i, x) \leq s(x_j , x)$ whenever $i>j$ to maintain the order in $D^*$ permutation.
% \paragraph{Scaling ICL Shot}
Limited by the maximum encoding length of current PLMs (e.g., 2048), the maximum $k$ of ICL is generally about 32~\citep{min-etal-2022-metaicl}. The upper bound of ICL when scaling up $k$ remains a question. 
% Recent works~\citep{von2022transformers, dai2022can} bridge the connection between ICL and gradient descent and understand the ICL as a kind of implicit finetuning. ICL involves one-pass editing of weighting parameters, while FT is able to iteratively update the parameters based on the full-size dataset. The gap between ICL and FT thus lies in update iterations and the consumed data. 
Intuitively, scaling up the shot number $k$ of ICL can further help ICL reach the capacity of finetuning. 
Beyond the maximum encoding length of 8192, further scaling up $k$ in an efficient way needs the incremental encoding technique.

% Although EVALM supports the maximum encoding length of 8192, further scaling up $k$ could exceed it, thus we need the incremental encoding technique to support the input length longer than training.
\paragraph{Incremental Encoding}
% random drop or online clustering or mean pooling over states of min-2 attention
% \fjt{1. key idea first, introduce incremental states (with notation possibly?); 2. how incremental states is updated when a new feature comes; 3. why eva save memory}
Incremental decoding~\citep{ott2019fairseq} enhances the sequence generation efficiency by caching useful historical states, namely \emph{incremental states}, for future usage, which saves memory from the redundant computation. 
Inspired by this, we devise incremental encoding, which updates EVA cache states incrementally, for long context encoding.
According to EVA architecture (\secref{sec:arch}), we maintain all local and remote features as incremental states $\mathcal{S}\text{: }\{{\mathbf{K}}^{l},{\mathbf{V}}^{l},\hat{\mathbf{K}}^{r},\hat{\mathbf{V}}^{r}\}$. When encoding the incoming tokens, we first concatenate them with local features and then compress full-chunk-sized local features into remote features and update $\mathcal{S}$, details in Algorithm~\ref{alg:incremental}.

% Given a sequence $\boldsymbol{s}$ with length of $L$, we obtain the chunked long sequence and compress them into remote features $\{\hat{\mathbf{K}}^r;\hat{\mathbf{V}}^r\}$ and the remaining as local features $\{{\mathbf{K}}^l;{\mathbf{V}}^l\}$. Based on this, when encoding the incoming tokens, we first concatenate them with local $\{{\mathbf{K}}^l;{\mathbf{V}}^l\}$ and then compress the part exceeding chunk size $c$ into remote $\{\hat{\mathbf{K}}^r;\hat{\mathbf{V}}^r\}$ and update them. 

% \begin{algorithm}
% \small
% \caption{Incremental Encoding}\label{alg:incremental}
% \KwIn{incoming $\mathbf{x}_t$, operation $\mathcal{M}(\cdot)$, chunk size $c$, incremental states $\mathcal{S}:\{{\mathbf{K}}^{l};{\mathbf{V}}^{l};\hat{\mathbf{K}}^{r};\hat{\mathbf{V}}^{r}\}$}
% \KwOut{$\text{EVA}(\mathbf{x}_t)$, updated $\mathcal{S}$}
% $\mathbf{q}_t$, $\mathbf{k}_t$, $\mathbf{v}_t = \mathrm{projection}(\mathbf{x}_t)$\;
% $\mathbf{K}^{l}=\mathrm{cat}([\mathbf{K}^{l}, \mathbf{k}_t])$, $\mathbf{V}^{l}=\mathrm{cat}([\mathbf{V}^{l}, \mathbf{v}_t])$\;
% \If{$\mathrm{length}(\mathbf{K}^{l})$ equals to $c$} {
% $\hat{\mathbf{K}}^{r}=\mathrm{cat}([\hat{\mathbf{K}}^{r},\mathcal{M}(\mathbf{K}^r)])$\;
% $\hat{\mathbf{V}}^{r}=\mathrm{cat}([\hat{\mathbf{V}}^{r},\mathcal{M}(\mathbf{V}^r)])$\;
% $\mathrm{empty}(\mathbf{K}^{l}), \mathrm{empty}(\mathbf{V}^{l})$\;
% }
% \KwRet{$\text{EVA}(\mathbf{x}_t)=\mathrm{softmax}(\mathbf{q}_t [\hat{\mathbf{K}}^r;\mathbf{K}^l]^\top)[\hat{\mathbf{V}}^r;\mathbf{V}^l]^\top$}
% \end{algorithm}

\begin{algorithm}
% \small
\caption{Incremental Encoding}\label{alg:incremental}
\begin{algorithmic}
\REQUIRE chunk size $c$, compression attention and pooling operation $\mathcal{M}(\cdot)$, incoming token $\mathbf{x}_t$, previous incremental states $\mathcal{S}_{t-1}:\{{\mathbf{K}}^{l}_{t-1},{\mathbf{V}}^{l}_{t-1},\hat{\mathbf{K}}^{r}_{t-1},\hat{\mathbf{V}}^{r}_{t-1}\}$
\ENSURE updated incremental states $\mathcal{S}_t$
\STATE $\mathbf{q}_t, \mathbf{k}_t, \mathbf{v}_t = \mathrm{projection}(\mathbf{x}_t)$
\STATE $\mathbf{K}^{l}_t=[\mathbf{K}^{l}_{t-1}; \mathbf{k}_t], \mathbf{V}^{l}_t=[\mathbf{V}^{l}_{t-1}; \mathbf{v}_t]$
\IF{$\mathrm{length}(\mathbf{K}_t^{l})$ is $c$}
  \STATE $\hat{\mathbf{K}}^{r}_t=[\hat{\mathbf{K}}^{r}_{t-1};\mathcal{M}(\mathbf{K}^l_t)], \hat{\mathbf{V}}^{r}_t=[\hat{\mathbf{V}}^{r}_{t-1};\mathcal{M}(\mathbf{V}^l_t)]$   
  \STATE $\mathbf{K}^{l}_t:=\emptyset, \mathbf{V}^{l}_t:=\emptyset$
\ELSE
  \STATE $\hat{\mathbf{K}}^{r}_t=\hat{\mathbf{K}}^{r}_{t-1},\hat{\mathbf{V}}^{r}_t=\hat{\mathbf{V}}^{r}_{t-1}$
\ENDIF
\STATE {\bfseries return} $\mathcal{S}_{t}:\{{\mathbf{K}}^{l}_{t},{\mathbf{V}}^{l}_{t},\hat{\mathbf{K}}^{r}_{t},\hat{\mathbf{V}}^{r}_{t}\}$
\end{algorithmic}

\end{algorithm}

Previously, encoding long-range context requires quadratic memory complexity, and incremental encoding consequently scales it down to linear by caching previous $\mathcal{S}$, where the memory consumption grows linearly along with the increase of $\mathcal{S}$. Powered by this, \textsc{EvaLM} further reduces the memory bottleneck and ensures the input length is scalable. In practice, the upper bound of encoding length is $32\times$ than training, and it is possible to encode an extremely long sequence into incremental states losslessly, with the compression rate $c$. Incremental encoding thus brings numerous benefits for many scenarios like ICL.
% \fjt{1. why incremental encoding is vital to in-context learning; 2. how to perform incremental encoding to in-context learning in practice}
For ICL, considering many test samples share the same demonstration sequence $\boldsymbol{s}^e$, we can encode it once, cache the long-term incremental states, and reuse them for further possible encoding. The test samples are then fed forward, conditioned on $\mathcal{S}$, to predict the result using \eqnref{eq:ppl-incre}. Reusing the incremental states of demonstration saves the extra overheads of scaling $k$.

% and it is mathematically equal to \eqnref{eq:ppl}.
% \begin{equation}
% \label{eq:ppl-incre}
% \begin{split}
%   P(Y_i|x)&=\sum_{1\leq t\leq L} \log p_t\left([D^*]^L\right)\textit{\small{~(one-pass)}} \\
%   + &\sum_{L<t\leq l}\log p_t\left([\mathcal{I}(x, Y_i)]^{l-L}\big\rvert D^*\right),
% \end{split}
% \end{equation}
% where the first term will be computed once, and the second term refers to the logits predicted by causal EVALM conditioned on the saved incremental states. Reusing the incremental states of demonstration saves the extra overheads of scaling $k$.

\section{Experiments}
In this section, we conduct in-context learning experiments to validate our \textsc{EvaLM} and its instruction-tuned version on various tasks.

\subsection{Experimental setting}
\label{exp:setup}
\paragraph{Pre-training}
We pre-trained \textsc{EvaLM}~(350M and 1.3B) on 32 NVIDIA A100 80G GPUs. 
The hyper-parameters of \textsc{EvaLM}s are identical to GPT3~\citep{brown2020gpt3} and OPT~\citep{zhang2022opt} in the same scale, where the hidden size, number of attention heads and number of layers are 1024, 16, 24 respectively for the 350M model and 2048, 32, 24 respectively for the 1.3B model.

\paragraph{Instruction Tuning} Following FLAN~\citep{FLAN2022Wei}, we experiment ICL on the downstream tasks using \textsc{EvaLM} instruction-tuned on FLAN datasets. FLAN dataset that belongs to the same cluster with the current test task is excluded during the instruction tuning stage, preventing the evaluation from data leakage and remaining our setting regarded as zero-shot or many-shot. There are three settings for instruction tuning in our experiments: a) IT: one-shot IT; b) MSIT: many-shot instruction tuning with a maximum of $8192$ per batch line; c) MSIT+\footnote{We are unable to apply MSIT+ on the 1.3B \textsc{EvaLM} due to memory limitations, we
leave it for the future model parallel version of \textsc{EvaLM}}: MSIT with a maximum of $2\times 8192$ per batch line. More instruction tuning details can be seen in Appendix \ref{appendix:IT}. 

\paragraph{In-Context Learning} 
We mainly follow~\citet{wu2022self} and~\citet{FLAN2022Wei} to select several datasets from different NLP tasks. We choose SST-2 and SST-5
 for sentiment classification~\citep{socher-etal-2013-recursive}, MNLI~\citep{williams-etal-2018-broad} for natural language inference, MultiRC~\citep{khashabi-etal-2018-looking} and BoolQ~\citep{clark-etal-2019-boolq} for reading comprehension, AgNews~\citep{zhang2015agnews} for topic classification, WSC~\citep{wsc2012hector} for coreference resolution, COPA~\citep{roemmele2011choice} for commonsense reasoning and Trec~\citep{trec2001ho} along with WiC~\citep{pilehvar-camacho-collados-2019-wic} for miscellaneous tasks.

We mainly adopt zero-shot and many-shot settings.
The \textit{zero-shot} setting directly wraps up the testing input with a task-specific template for inference.
The \textit{many-shot} approach randomly selects $k$ demonstrations from the training set and uses the same demonstrations for the whole test set. This approach is universally used as dataset-level ICL. We also adopt Top-$k$ approach following~\citeauthor{wu2022self} in~\secref{sec:discuss}. Prompt designs are detailed in Appendix~\ref{appendix:prompt}.

We find the best shot number on the validation set  
 and test on the test set when the label of the test set is available (AgNews, Trec, SST-5). For other datasets, we split 500 samples from each training set as a validation set and report our results on the test set. The demonstration number $k$ is set from 1 to 2000, please refer to Appendix~\ref{appendix:icl details} for more in-context learning details.

\paragraph{Baselines} We use OPT~\citep{zhang2022opt} as the main baseline due to its similar model architecture, number of parameters, training flops, training data, and training framework to our \textsc{EvaLM}, allowing for a fair comparison. 
We conduct experiments using models of 350M and 1.3B parameters.

\subsection{Main Results}
The overall in-context learning results are shown in \tabref{tab:icl}.
Based on this, we make the following observations.

\paragraph{Scaling up demonstration examples helps ICL} Since \textsc{EvaLM} pre-trained with longer sequence length and adapted for extrapolation, we can use more demonstrations when conducting in-context learning experiments. At both 350M and 1.3B scale, \textsc{EvaLM} outperforms OPT on both zero-shot and many-shot settings, and tends to achieve the best score at higher average shot number $k$ (about 10 times to OPT). This shows that long-range \textsc{EvaLM} can effectively utilize the information in demonstrations to get better results. The specific best shot numbers for each dataset and model are in Appendix~\ref{appendix:icl details}.

\paragraph{MSIT arouses the potential of many-shot ICL}
\tabref{tab:icl} shows that the model with MSIT, especially MSIT+, obtains the most growth, from zero-shot to many-shot setting, which is indicated by the relative improvement scores. This is partly because MSIT learns to align the language modeling with many-shot in-context learning scenarios, making it more suitable for testing in many-shot settings. Another reason is the relatively poor zero-shot performance with MSIT. A potential explanation is that learning too many tasks fills the capacity of small PLMs, which can be harmful to their zero-shot performance, as mentioned in FLAN~\citep{FLAN2022Wei}. Thus, we speculate that combining MSIT and scaling in-context shot number $k$ together is essential for getting the best in-context learning results.

\paragraph{Larger PLMs suit many-shot ICL}
Both \textsc{EvaLM}-1.3B and OPT-1.3B show more significant progress compared with the 350M model. This is also consistent with the rule of scaling law~\citep{chung_scaling_2022}. Large PLM contains more knowledge and can better conduct ICL through more demonstrations. This suggests that scaling up ICL may yield greater benefits on larger models.

% \sansa{(Conclusion: without IT, \textsc{EvaLM} achieves better performance than OPT and tends to achieve the best score at higher $k$)}

\begin{table*}[t]
\small
\setlength\tabcolsep{2.3pt}
    \centering
    \caption{Main results of in-context learning on diverse tasks. The light grey shade refers to the ablation modules of IT. We average the shot number of demonstrations when the best score is achieved. The best overall results are bolded. The abbreviation avg. is for average, imprv. is for improvement, acc is for accuracy. \textbf{The relative improvements of models in the many-shot setting are compared with the same model but in the zero-shot setting respectively.}}
    \begin{tabular}{l|cc|c|cc|cc|c|c|c|ccc}
    \toprule
    \multirow{2}{*}{\textbf{Models}} & \multicolumn{2}{c|}{Sentiment} & \multicolumn{1}{c|}{NLI}  & \multicolumn{2}{c|}{Miscellaneous}   & \multicolumn{2}{c|}{Reading}  & \multicolumn{1}{c|}{Topic} & \multicolumn{1}{c|}{Coreference}  &\multicolumn{1}{c|}{Commonsense} & \multirow{2}{*}{\tabincell{c}{\textbf{Avg.}\\\textbf{acc}}}&\multirow{2}{*}{\textbf{Imprv.}} &\multirow{2}{*}{\tabincell{c}{\textbf{Avg.}\\\textbf{shot}}}\\  
    & SST-2  & SST-5   & MNLI    & ~~Trec & WiC & MultiRC & BoolQ & AgNews  & WSC    & COPA & \\
    \midrule
    \multicolumn{14}{c}{\textbf{\textit{zero-shot}}} \\
    \midrule
     OPT-350M & 64.6 & 29.9  & 21.6    & 23.0  & 52.7 &46.3 &53.8 &50.9 &63.4 & 65.0 & 47.1 & - & -\\ 
         \textsc{EvaLM}-350M & 61.4 & 25.8  & 27.5  & 21.8 & 51.7 & 56.9 & 56.9 & 46.6 & 63.5 & 64.0 & 47.6 & - &-\\
         \rowcolor{shadecolor}\quad w/ MSIT  & 50.8 & 28.2  & 28.6 &  20.4 & 50.6 & 43.3 & 49.9 & 47.5 & 63.5 & 65.0 & 44.8 & - &- \\
         \rowcolor{shadecolor}\quad w/ MSIT+ & 64.0 & 29.3  & 28.0   & 22.2 & 50.4 & 42.0 & 53.3 & 48.6 & 63.5 & 62.0 & 46.3 & - & -\\
         OPT-1.3B & 73.0 & 31.3  & 20.0  & 22.0 & 50.3 & 41.7 &51.4  &56.6  &62.5 &72.0  & 48.1 & -&- \\ 
     \textsc{EvaLM}-1.3B & 82.3 & 31.3  & 21.6  & 22.8 & 52.1 & 41.7 &58.5  &55.3 &58.6  &72.0 & 49.6 & -&-\\ 
     \rowcolor{shadecolor}\quad w/ MSIT & 58.4 & 33.5  & 21.6  & 23.0 & 52.3 & 52.2 & 52.4 & 55.2 & 58.2 & 71.0 & 47.8 & -&-\\
    
    \midrule
    \multicolumn{14}{c}{\textbf{\textit{many-shot}}} \\
    \midrule
    OPT-350M & 62.3  & 31.0  & 33.8   & 27.6  & 51.6  & 57.2  & 62.8  & 63.8  & 63.4  & 64.0  & 51.7 & 4.6 &10\\ 
         \textsc{EvaLM}-350M & 61.0  & 32.3   & 32.1   & \textbf{49.6}  & 52.0 & 55.7  & 60.6  & 69.7  & 63.4  & 63.0  & 53.9 & 6.3 &97\\ 
         \rowcolor{shadecolor}\quad w/ MSIT & 65.2  & 31.2   & 34.1   & 39.4  & 52.4 & 53.4  & 57.5  & 70.1  & 63.5  & 72.0  & 53.9 & 9.1 &236\\
         \rowcolor{shadecolor}\quad w/ MSIT+ & 70.6  & 33.7  & \textbf{34.5}    & 40.4  & 50.4 & 53.1  & 59.2  & \textbf{73.3}  & 63.5  & 73.0 & 55.2 & 8.8 &208\\
         OPT-1.3B & 73.0  & 40.1  & 31.3   & 45.6  & 50.3  & 52.5  & 65.2  & 60.0  & 63.4  & \textbf{74.0}  & 55.3 & 7.3 &14\\ 
         \textsc{EvaLM}-1.3B & 76.6  & 40.4   & 30.2   & 46.8  & \textbf{54.3}  & 58.9  & 62.5 & 62.5 & 63.5  & \textbf{74.0}  & 57.0 &7.3 &152\\
         \rowcolor{shadecolor}\quad w/ MSIT & \textbf{84.2}  & \textbf{45.4}  & 33.9   & 49.4  & 54.2  & \textbf{60.2}  & \textbf{64.2} & 63.2 & \textbf{65.4}  & \textbf{74.0}  & \textbf{59.4} & \textbf{11.6}&269\\

    \bottomrule
    \end{tabular}

    \label{tab:icl}
\end{table*}

\begin{table}[htb]
\setlength\tabcolsep{2pt}
 \centering
 % \small
 \caption{Average accuracy and input length when achieving highest scores over all datasets with different IT strategies.}
    \begin{threeparttable}
    \begin{tabular}{c|cccccc}
        \toprule
          \multirow{2}{*}{\textbf{Models}} & \multicolumn{2}{c}{\textbf{Vanilla}} & \multicolumn{2}{c}{\textbf{w/ IT}} & \multicolumn{2}{c}{\textbf{w/ MSIT}}  \\  
          &\textbf{Acc.} & \textbf{Length} & \textbf{Acc.} & \textbf{Length}  & \textbf{Acc.} & \textbf{Length} \\  
        \midrule
          OPT-350M &51.7  &584.5  &50.9  &560.2  &51.5 &1592.3 \\
          \textsc{EvaLM}-350M & 53.9 &3904.9  &53.7  &3682  &53.9 &8087.7\\
        \midrule
          OPT-1.3B &55.3 &665.0 &54.5 & 670.3 &54.8 &1809.6  \\
          \textsc{EvaLM}-1.3B &  57.0 &7337.0  &56.9 &8140.6 &59.4 &12558.0 \\
        \bottomrule
    \end{tabular}
    \end{threeparttable}
    
    \label{tab: instruction tuning}
\end{table}

\subsection{Analysis on MSIT}
\paragraph{Efficacy of MSIT} To further investigate the effectiveness of MSIT, we average the best ICL results on instruction-tuned \textsc{EvaLM} as the shot number increases. 
Considering the average example length of different datasets varying from each other, we also count the length of demonstrations at the peak of accuracy instead of using the shot number. 
All results are averaged over 10 datasets in \tabref{tab: instruction tuning}. We also conduct the same experiment on the same size OPT model, but with 2048 tokens per batch line for MSIT. 

We observe that as the number of IT examples grows, the average lengths of many-shot examples increase accordingly, for both OPT and \textsc{EvaLM}.
It reflects that MSIT indeed learns the alignment with many-shot ICL.
Such alignment helps \textsc{EvaLM}'s enhance its capability on many-shot ICL, but becomes helpless or even harmful to OPT.
A possible reason is that \textsc{EvaLM} is specialized in long-range language modeling with extrapolation whilst OPT is not.
Thus, many-shot ICL is more suitable to \textsc{EvaLM} with MSIT.

\begin{figure}[t]
\centering
\includegraphics[height=3.1cm]{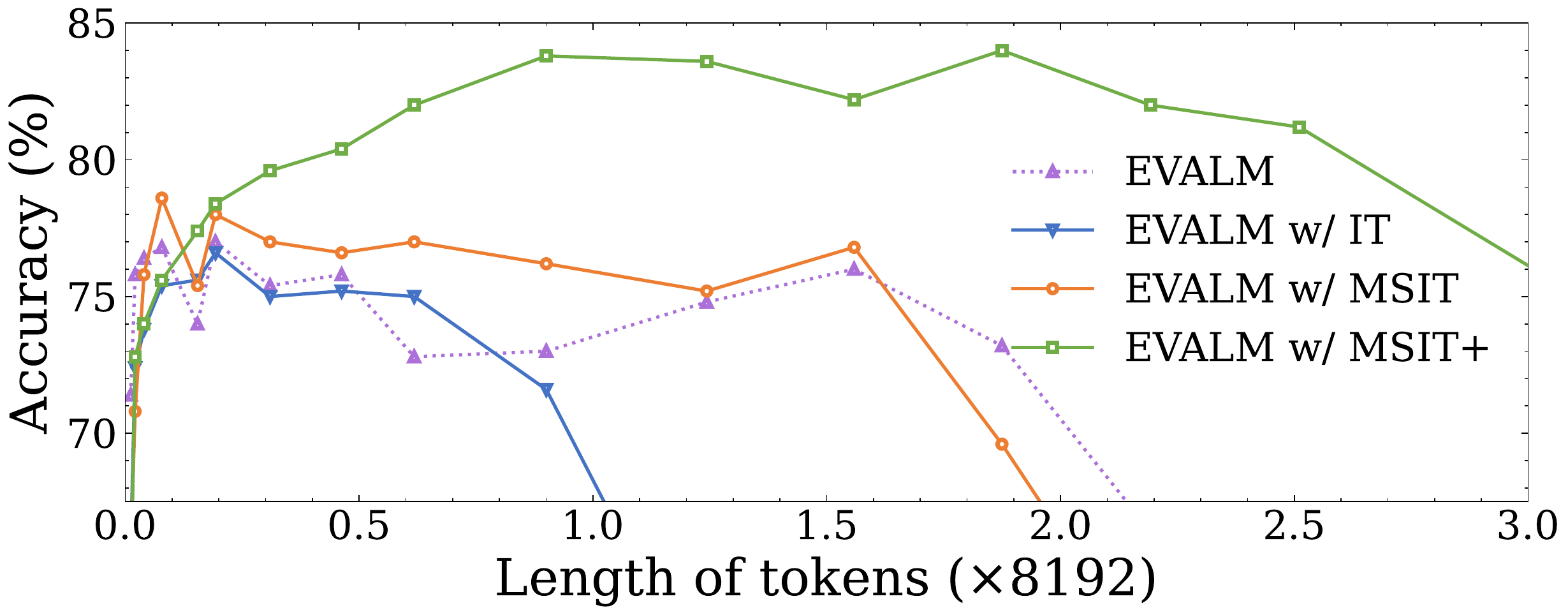}
\caption{The ICL accuracy curve along with demonstration length on Trec dataset using the Top-$k$ approach, for \textsc{EvaLM}-350M models with different instruction tuning strategies.}
\label{fig:ITshot}
\end{figure}

\begin{figure}[t]
\centering
\includegraphics[height=3cm]{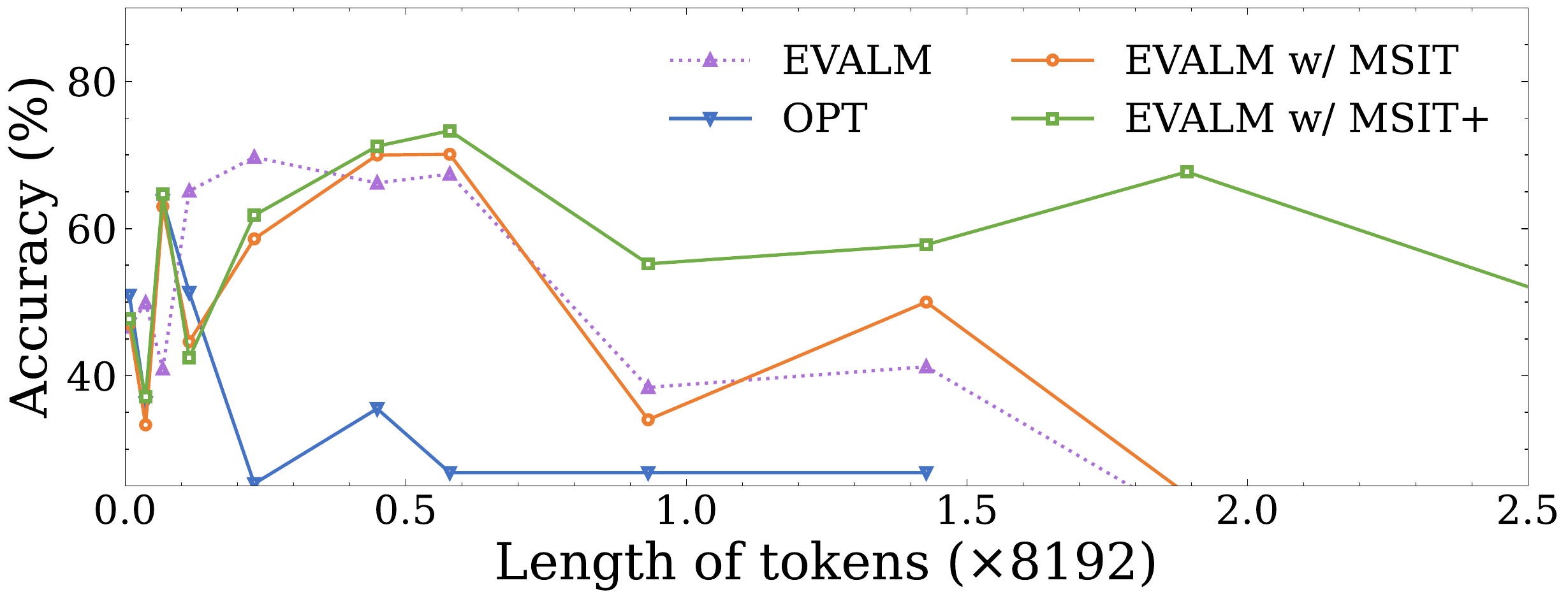}
\caption{The ICL accuracy curve along with demonstration length on AgNews dataset using the random in-context examples, for \textsc{EvaLM}-350M with different instruction tuning strategies and OPT-350M.}
\label{fig:ITshot-ag}
\end{figure}

\paragraph{Scaling $k$-shot ICL}
We dig into the specific accuracy curve as the demonstration length rises, taking the AgNews dataset using randomly selected many-shot ICL and the Trec dataset using Top-$k$ many-shot ICL as examples. Please refer to~\secref{sec:discuss} for more analysis about the Top-$k$ setting. We choose this setting to analyze considering the robustness of the approach and stability of the curve. As shown in~\figref{fig:ITshot}, we observe that \textsc{EvaLM} without instruction tuning or just with one-shot instruction tuning achieves the highest accuracy within $128$ shots, which corresponds to around $2$k length, and further adding demonstrations makes the accuracy curve drop quickly. 
With many-shot instruction tuning, the best accuracy is improved and the heavy drop gets alleviated. 
With instruction tuning on longer range~(MSIT+), the accuracy grows steadily along with increasing demonstration length and peaks at $768$ shots, which corresponds to around $15$k length. Similar trends can be found in~\figref{fig:ITshot-ag} but OPT can not.
This trend indicates that MSIT encourages our language model to achieve higher accuracy, and scaling up the shot number of IT further improves the upper bound of scaling in-context learning on downstream tasks.

However, the increasing trend is not endless, even for MSIT+. When the length of demonstration examples reaches $20$k, the rapid drop of the accuracy curve can be seen. The possible reasons are listed as follows. 
On the one hand, modeling the longer input length relies on the extrapolation ability of models, and the size of models could also be the limiting factor. 
There are more discussions in~\secref{sec:discuss}. 
On the other hand, the setting of enlarging the demonstration example sizes, in both of instruction tuning and in-context learning, is under-explored, due to the lack of pre-trained LRLMs. 
Therefore, advanced ICL algorithms are demanded for further investigation on many-shot in-context learning.

% \paragraph{Document  understanding}
% scrolls benchmark

\subsection{Discussion}
\label{sec:discuss}
% \paragraph{Language Modeling Ability}
% We follow \citep{zhang_opt_2022,Dao_2022_hungry} to use GPT-3~\citep{brown2020language} style prompt to show a common NLU ability of these large pretraining model in the same size in Table \ref{tb:super_glue}
% \sansa{(1, Zero-shot superglue to show basic language modeling ability of \textsc{EvaLM} \tabref{tb:super_glue}; move to appendix; maybe still need to compare with H3}
\paragraph{Extrapolation Ability}
To ensure the extrapolation ability of \textsc{EvaLM}, we simply adopt CPE (\secref{sec:arch}), and incremental encoding (\secref{sec:icl}) is deployed to save memory consumption. 
With these techniques, \textsc{EvaLM} is able to encode super-long inputs, i.e. $256$k on 80G NVIDIA A100, during inference. 
% From \figref{fig:inf memo} we can see that given the 80G memory limitation of NVIDIA A100, the maximum supported input length is $256$k. 
For comparison, we also adapt the incremental encoding technique to the OPT model of the same size, whose maximum context size still lags behind \textsc{EvaLM}'s. 
Detailed comparison of memory consumption between OPT and \textsc{EvaLM} can be found in Appendix~\ref{appendix:icl details}.

\begin{figure}[b]
\centering
\includegraphics[height=3cm]{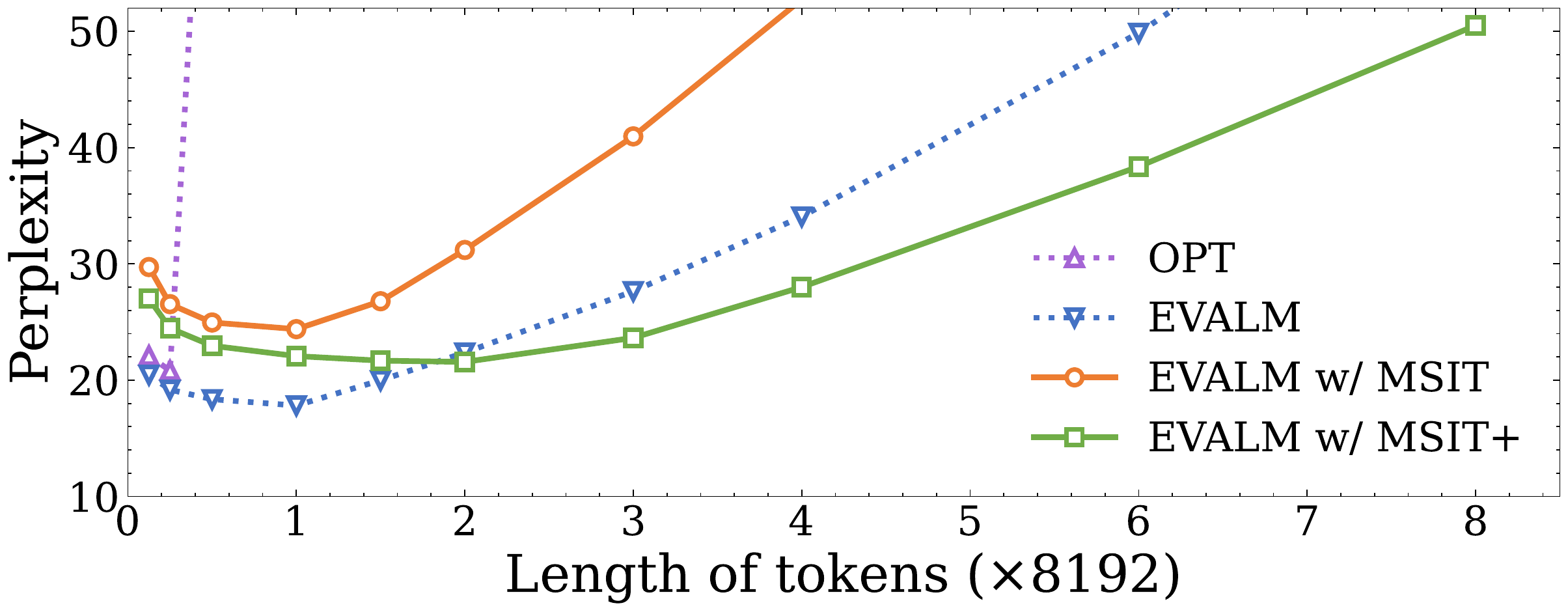}
\caption{The perplexity of OPT-350M and \textsc{EvaLM}-350M on PG-19 dataset when the length of input sequence scaled up. The extrapolation of OPT starts from 2048 and others from 8192}
\label{fig:extrapolation}
\end{figure}

Based on this, we further evaluate the extrapolation ability of different models using perplexity. 
The experiment is conducted on PG-19 dataset~\citep{raecompressive2019}, a dataset focusing on long-range language modeling, following the setting by \citet{zhang2022cab}.
The perplexity curve along with the input length is addressed in~\figref{fig:extrapolation}. The perplexity of OPT grows steeply once the input length is over 2048, indicating its poor extrapolation ability. The vanilla \textsc{EvaLM} with MSIT increases the perplexity, which is expected considering that the instruction tuning will adapt PLMs from the general corpus towards several specific tasks. Compared with MSIT, \textsc{EvaLM} with MSIT+ achieves lower perplexity even lower than the vanilla model. MSIT+ also reaches the lowest perplexity at a larger input length around $16$k. These observations explain why the OPT benefits less while \textsc{EvaLM} benefits more from MSIT especially MSIT+ in~\tabref{tab: instruction tuning}, \figref{fig:ITshot} and \figref{fig:ITshot-ag}.

\paragraph{Top-$k$ ICL}
Following~\citeauthor{wu2022self}, we also deploy \textit{Top-$k$} approach (instance-level) which selects the $k$ most similar samples from the training dataset based on embedding similarities~\citep{liu2022makes, gao2021making} and puts the samples with higher similarity closer to the testing input. We conduct Top-$k$ ICL in our commonly used datasets to further verify the effects of MSIT. 

\begin{table}[t]
\setlength\tabcolsep{2.5pt}
 \centering
 \small
 \caption{Results of using Top-$k$ ICL approach. The light grey shade refers to the ablation modules of IT. The best results are bolded. The abbreviation avg. is for average.}
    \begin{threeparttable}
    \begin{tabular}{l|ccccc|c}
        \toprule
          \textbf{Models} & SST-2 & SST-5 & MNLI & Trec  & AgNews &\textbf{Avg.} \\  
          \midrule
    OPT-350M & 86.1 & 44.5 & \textbf{33.8} & 74.8  & 91.0 & 66.0\\ 
     \textsc{EvaLM}-350M & 88.2 & 46.5 & 29.5 & 76.8 & 91.8 & 66.6\\
     \rowcolor{shadecolor}\quad w/ MSIT & 86.0 & 43.6 & 27.2 & 78.0 & 90.9 & 65.1\\
         \rowcolor{shadecolor}\quad w/ MSIT+  & \textbf{88.3} & 44.9 & 27.7 & \textbf{83.8} & \textbf{91.9} & \textbf{67.3}\\
     OPT-1.3B & 86.7 & 43.2 & 25.1 & 77.0 & 91.3 & 64.7\\
     \textsc{EvaLM}-1.3B  & 87.7 & \textbf{47.4} & 30.2 &79.0& 91.8 & 67.2\\ 
     \rowcolor{shadecolor}\quad w/ MSIT & 88.2 & 47.0 & 32.2 & 76.0 & 91.0 & 66.9 \\
        \bottomrule
    \end{tabular}
    \end{threeparttable}
    
    \label{tab:topk}
\end{table}

As shown in~\tabref{tab:topk}, the advantage of MSIT is not much significant as using a randomly selected \textit{many-shot} ICL approach in~\tabref{tab:icl}. 
There are two possible aspects to explain this. Firstly, \textsc{EvaLM} with MSIT aims to establish a robust PLM to close the gap of downstream task performances by randomly selected or carefully picked in-context examples, whereas Top-$k$ is a competitive and robust in-context examples selector to each instance regardless of PLMs. Such consistent and overlapped goal with Top-$k$ algorithm prevents \textsc{EvaLM} from gaining further improvements. Second, the permutation of demonstration examples in Top-$k$ is not in the same pattern as our instruction tuning, where the latter is randomly ordered instead of sorted by similarity. Even in this situation, MSIT+ still shows a positive effect on most of the tasks, showing the considerable effects of more shots instruction tuning.

Besides, Top-$k$ approach, an instance-level ICL algorithm, selects different demonstration examples for each test sample, demanding heavy computation resources in ICL inference.
In contrast, the random approach, a dataset-level ICL algorithm, is much cheaper by sharing and caching incremental encoded examples for all the test samples.
Thus, we believe that the random approach or advanced dataset-level ICL algorithms are more compatible and promising to LRLMs.

\paragraph{Efficiency}

We test the efficiency of \textsc{EvaLM} with training FLOPs and inference times, which are considered crucial for PLMs in upstream training and downstream usage.
As shown in~\tabref{tab:training-efficiency}, \textsc{EvaLM} can achieve better in-context learning performance with OPT in the same size with even lower training costs. This is due to the efficiency of the causal EVA and our deduplicated training data. Compared with pre-training, the cost of instruction tuning is significantly lower, making it a more easily adopted way to improve the in-context learning performance of PLMs.

\begin{table}[ht]
\setlength\tabcolsep{3pt}
 \centering
 % \small
 \caption{Training FLOPs of different models}
    \begin{threeparttable}
    \begin{tabular}{c|cc}
        \toprule
          \textbf{Models} & Vanilla   & w/ MSIT  \\  
        \midrule
          OPT-350M &3.84E+20    &1.60E+18 \\
          \textsc{EvaLM}-350M &2.80E+20   &1.15E+18\\
        \midrule
          OPT-1.3B &1.42E+21 &5.94E+18 \\
          \textsc{EvaLM}-1.3B &1.03E+21  &4.27E+18 \\
        \bottomrule
    \end{tabular}
    \end{threeparttable}
    
    \label{tab:training-efficiency}
\end{table}

\begin{figure}[thb]
\centering
\includegraphics[height=3cm]{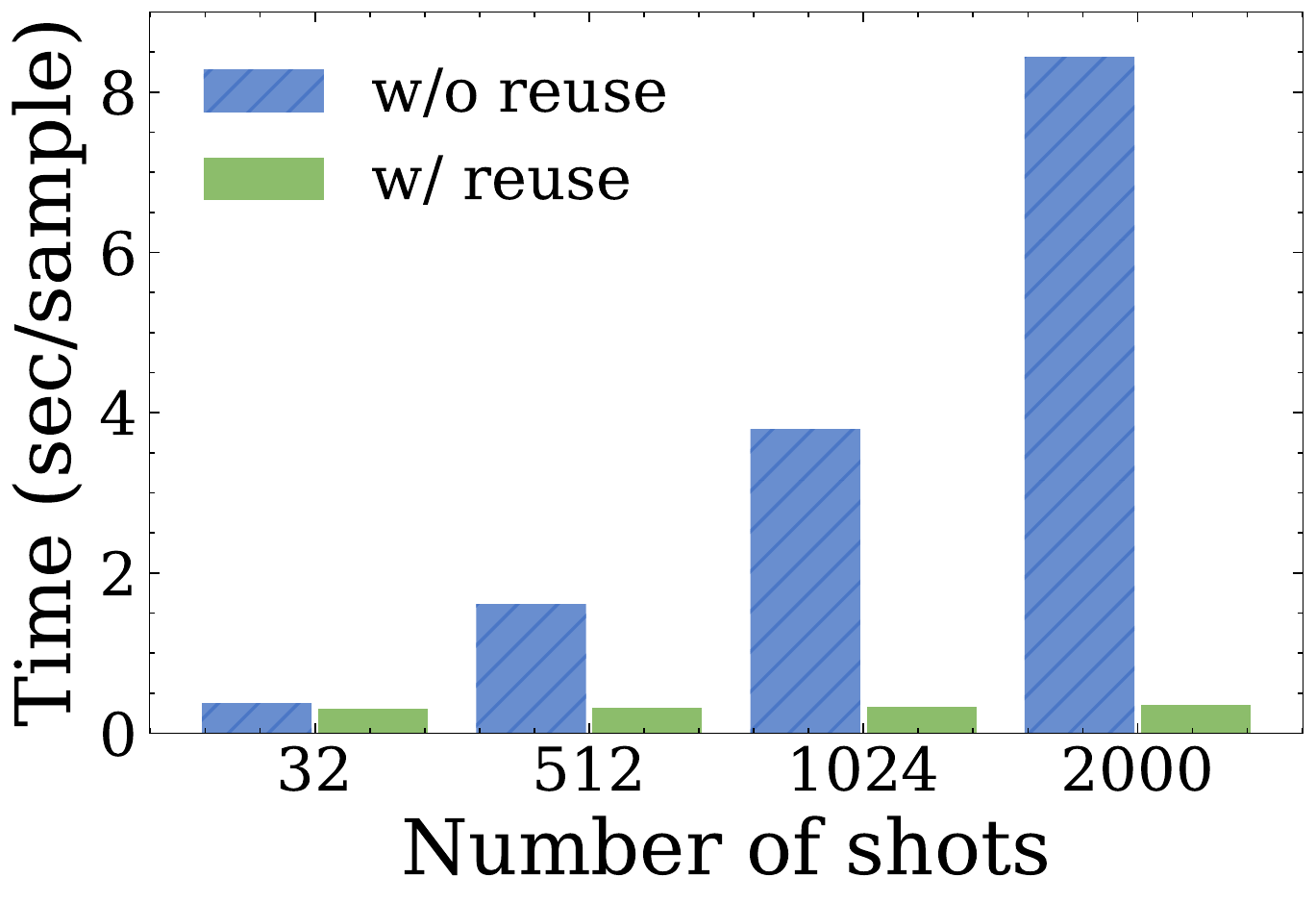}
\caption{Inference time for each sample on SST-5 with or without the reuse of incremental states. 2000 shots corresponds to $58$k input length for this dataset}
\label{fig:inf speed}
\end{figure}

As for inference efficiency, according to \secref{sec:icl}, with incremental encoding and the reuse of incremental states, the additional cost of $k$-shot ICL when scaling up $k$ in \textsc{EvaLM} is relatively low in many-shot settings. \figref{fig:inf speed} illustrates the time consumption of \textsc{EvaLM}-350M for each test sample along with the number of shots $k$, and the results are conducted on SST-5 averaged over $1000$ samples. It indicates that without the reuse of incremental states, the overheads grow rapidly while reusing saves redundant computation. The consumption of first encoding long-range demonstrations is diluted by the number of test samples. 

\section{Conclusions \& Future Work}
The under-investigated pre-trained long-range language model limits the exploration of more shots instruction tuning and in-context learning. In this work, we first pre-train a casual language model \textsc{EvaLM} based on an efficient attention mechanism EVA, successfully enabling training with $8$k tokens and extrapolating with $256$k-length contexts. With techniques such as incremental encoding for efficiency and circular position embedding for extrapolation, we consequently inspect the effectiveness of increasing the shot number of both instruction tuning and in-context learning using \textsc{EvaLM}. Experimental results across a variety of tasks show \textsc{EvaLM} with many-shot instruction tuning and plus outperforms the same size OPT by 4.1\% accuracy on average. Interestingly, we find that many-shot instruction tuning can help ICL achieve higher performance with larger demonstrations, and with longer instructions, this phenomenon is more obvious. Notably, such many-shot ICL, with incremental encoding and caching, demands rare extra computational overheads.

\textsc{EvaLM} takes the first step towards many-shot in-context learning with pre-trained long-range language models, but it still has several limitations.
First, due to our limited computational resources, the experimented \textsc{EvaLM} is relatively small in model size compared to existing large-scale language models, e.g. GPT, OPT and PaLM.
We will actively work on scaling up its capacity, and it would be interesting to expect its performance on larger LRLMs.
Second, although the backbone attention model EVA is efficient and competitive with vanilla attention, it still struggles to scale to longer sequence modeling, due to its quadratic complexity to sequence length in causal language modeling.
We will improve LRLMs with linear attention mechanisms to further scale up the reachable length of contexts.
% Second, the performance gain from many-shot in-context learning becomes marginal as the in-context examples grows.
% It is worth exploration on many-shot in-context learning methods to make the performance gain grow consistently and steadily to the size of in-context sample.
% Third, although \textsc{EvaLM} is fledged with incremental encoding, it is still slow for instance-level in-context learning, while instance-level one is shown more superior to dataset-level one~\citep{wu2022self}.
% Thus we believe that advanced dataset-level in-context learning with LRLM is worth exploring for its high efficiency by reusing incremental states.
% Besides, fast instance-level in-context learning on LRLM is also a challenging but promising direction.
Third, when scaling up in-context examples, \textsc{EvaLM} is incapable of gaining performance from marginal ones, consistently.
We will explore new many-shot in-context learning algorithms that consistently gain performance from the increasing number of in-context examples.

\section{Acknowledgements}
We thank Lin Zheng for proposing the state-of-art efficient attention EVA and providing a well-designed codebase. This work is partially supported by the Shanghai Committee of Science and Technology (Grant No. 21DZ1100100) and the joint research scheme of the National Natural Science Foundation of China (NSFC) and the Research Grants Council (RGC) under grant number N\_HKU714/21.

\bibliography{custom}

\begin{thebibliography}{62}
\providecommand{\natexlab}[1]{#1}
\providecommand{\url}[1]{\texttt{#1}}
\expandafter\ifx\csname urlstyle\endcsname\relax
  \providecommand{\doi}[1]{doi: #1}\else
  \providecommand{\doi}{doi: \begingroup \urlstyle{rm}\Url}\fi

\bibitem[Ainslie et~al.(2020)Ainslie, Ontanon, Alberti, Cvicek, Fisher, Pham,
  Ravula, Sanghai, Wang, and Yang]{ainslie2020etc}
Ainslie, J., Ontanon, S., Alberti, C., Cvicek, V., Fisher, Z., Pham, P.,
  Ravula, A., Sanghai, S., Wang, Q., and Yang, L.
\newblock {ETC}: Encoding long and structured inputs in transformers.
\newblock In \emph{Proceedings of the 2020 Conference on Empirical Methods in
  Natural Language Processing (EMNLP)}, pp.\  268--284, Online, 2020.
  Association for Computational Linguistics.

\bibitem[Anonymous(2023)]{anonymous2023efficient}
Anonymous.
\newblock Efficient attention via control variates.
\newblock In \emph{Submitted to The Eleventh International Conference on
  Learning Representations}, 2023.
\newblock under review.

\bibitem[Beltagy et~al.(2020)Beltagy, Peters, and Cohan]{beltagy2020Longformer}
Beltagy, I., Peters, M.~E., and Cohan, A.
\newblock Longformer: The long-document transformer.
\newblock \emph{ArXiv preprint}, abs/2004.05150, 2020.

\bibitem[Black et~al.(2022)Black, Biderman, Hallahan, Anthony, Gao, Golding,
  He, Leahy, McDonell, Phang, Pieler, Prashanth, Purohit, Reynolds, Tow, Wang,
  and Weinbach]{black_gpt-neox-20b_2022}
Black, S., Biderman, S., Hallahan, E., Anthony, Q., Gao, L., Golding, L., He,
  H., Leahy, C., McDonell, K., Phang, J., Pieler, M., Prashanth, U.~S.,
  Purohit, S., Reynolds, L., Tow, J., Wang, B., and Weinbach, S.
\newblock {GPT}-{N}eo{X}-20{B}: An open-source autoregressive language model.
\newblock In \emph{Proceedings of BigScience Episode {\#}5 -- Workshop on
  Challenges {\&} Perspectives in Creating Large Language Models}, pp.\
  95--136, virtual+Dublin, 2022. Association for Computational Linguistics.

\bibitem[Brown et~al.(2020{\natexlab{a}})Brown, Mann, Ryder, Subbiah, Kaplan,
  Dhariwal, Neelakantan, Shyam, Sastry, Askell, Agarwal, Herbert{-}Voss,
  Krueger, Henighan, Child, Ramesh, Ziegler, Wu, Winter, Hesse, Chen, Sigler,
  Litwin, Gray, Chess, Clark, Berner, McCandlish, Radford, Sutskever, and
  Amodei]{brown2020gpt3}
Brown, T.~B., Mann, B., Ryder, N., Subbiah, M., Kaplan, J., Dhariwal, P.,
  Neelakantan, A., Shyam, P., Sastry, G., Askell, A., Agarwal, S.,
  Herbert{-}Voss, A., Krueger, G., Henighan, T., Child, R., Ramesh, A.,
  Ziegler, D.~M., Wu, J., Winter, C., Hesse, C., Chen, M., Sigler, E., Litwin,
  M., Gray, S., Chess, B., Clark, J., Berner, C., McCandlish, S., Radford, A.,
  Sutskever, I., and Amodei, D.
\newblock Language models are few-shot learners.
\newblock In Larochelle, H., Ranzato, M., Hadsell, R., Balcan, M., and Lin, H.
  (eds.), \emph{Advances in Neural Information Processing Systems 33: Annual
  Conference on Neural Information Processing Systems 2020, NeurIPS 2020,
  December 6-12, 2020, virtual}, 2020{\natexlab{a}}.

\bibitem[Brown et~al.(2020{\natexlab{b}})Brown, Mann, Ryder, Subbiah, Kaplan,
  Dhariwal, Neelakantan, Shyam, Sastry, Askell, Agarwal, Herbert{-}Voss,
  Krueger, Henighan, Child, Ramesh, Ziegler, Wu, Winter, Hesse, Chen, Sigler,
  Litwin, Gray, Chess, Clark, Berner, McCandlish, Radford, Sutskever, and
  Amodei]{brown2020language}
Brown, T.~B., Mann, B., Ryder, N., Subbiah, M., Kaplan, J., Dhariwal, P.,
  Neelakantan, A., Shyam, P., Sastry, G., Askell, A., Agarwal, S.,
  Herbert{-}Voss, A., Krueger, G., Henighan, T., Child, R., Ramesh, A.,
  Ziegler, D.~M., Wu, J., Winter, C., Hesse, C., Chen, M., Sigler, E., Litwin,
  M., Gray, S., Chess, B., Clark, J., Berner, C., McCandlish, S., Radford, A.,
  Sutskever, I., and Amodei, D.
\newblock Language models are few-shot learners.
\newblock In Larochelle, H., Ranzato, M., Hadsell, R., Balcan, M., and Lin, H.
  (eds.), \emph{Advances in Neural Information Processing Systems 33: Annual
  Conference on Neural Information Processing Systems 2020, NeurIPS 2020,
  December 6-12, 2020, virtual}, 2020{\natexlab{b}}.

\bibitem[Chen et~al.(2021)Chen, Zeng, Ji, and Yang]{chen2021skyformer}
Chen, Y., Zeng, Q., Ji, H., and Yang, Y.
\newblock Skyformer: Remodel self-attention with gaussian kernel and
  nystr{\textbackslash}''om method.
\newblock In Beygelzimer, A., Dauphin, Y., Liang, P., and Vaughan, J.~W.
  (eds.), \emph{Advances in Neural Information Processing Systems}, 2021.

\bibitem[Choi et~al.(2022)Choi, Jo, Jang, and Seo]{choi2022prompt}
Choi, E., Jo, Y., Jang, J., and Seo, M.
\newblock Prompt injection: Parameterization of fixed inputs.
\newblock \emph{ArXiv preprint}, abs/2206.11349, 2022.

\bibitem[Choromanski et~al.(2021)Choromanski, Likhosherstov, Dohan, Song, Gane,
  Sarl{\'{o}}s, Hawkins, Davis, Mohiuddin, Kaiser, Belanger, Colwell, and
  Weller]{choromanski2021rethinking}
Choromanski, K.~M., Likhosherstov, V., Dohan, D., Song, X., Gane, A.,
  Sarl{\'{o}}s, T., Hawkins, P., Davis, J.~Q., Mohiuddin, A., Kaiser, L.,
  Belanger, D.~B., Colwell, L.~J., and Weller, A.
\newblock Rethinking attention with performers.
\newblock In \emph{9th International Conference on Learning Representations,
  {ICLR} 2021, Virtual Event, Austria, May 3-7, 2021}. OpenReview.net, 2021.

\bibitem[Chowdhery et~al.(2022)Chowdhery, Narang, Devlin, Bosma, Mishra,
  Roberts, Barham, Chung, Sutton, Gehrmann, et~al.]{chowdhery2022palm}
Chowdhery, A., Narang, S., Devlin, J., Bosma, M., Mishra, G., Roberts, A.,
  Barham, P., Chung, H.~W., Sutton, C., Gehrmann, S., et~al.
\newblock Palm: Scaling language modeling with pathways.
\newblock \emph{ArXiv preprint}, abs/2204.02311, 2022.

\bibitem[Chung et~al.(2022)Chung, Hou, Longpre, Zoph, Tay, Fedus, Li, Wang,
  Dehghani, Brahma, Webson, Gu, Dai, Suzgun, Chen, Chowdhery, Narang, Mishra,
  Yu, Zhao, Huang, Dai, Yu, Petrov, Chi, Dean, Devlin, Roberts, Zhou, Le, and
  Wei]{chung_scaling_2022}
Chung, H.~W., Hou, L., Longpre, S., Zoph, B., Tay, Y., Fedus, W., Li, E., Wang,
  X., Dehghani, M., Brahma, S., Webson, A., Gu, S.~S., Dai, Z., Suzgun, M.,
  Chen, X., Chowdhery, A., Narang, S., Mishra, G., Yu, A., Zhao, V., Huang, Y.,
  Dai, A., Yu, H., Petrov, S., Chi, E.~H., Dean, J., Devlin, J., Roberts, A.,
  Zhou, D., Le, Q.~V., and Wei, J.
\newblock Scaling instruction-finetuned language models.
\newblock \emph{ArXiv preprint}, abs/2210.11416, 2022.

\bibitem[Clark et~al.(2019)Clark, Lee, Chang, Kwiatkowski, Collins, and
  Toutanova]{clark-etal-2019-boolq}
Clark, C., Lee, K., Chang, M.-W., Kwiatkowski, T., Collins, M., and Toutanova,
  K.
\newblock {B}ool{Q}: Exploring the surprising difficulty of natural yes/no
  questions.
\newblock In \emph{Proceedings of the 2019 Conference of the North {A}merican
  Chapter of the Association for Computational Linguistics: Human Language
  Technologies, Volume 1 (Long and Short Papers)}, pp.\  2924--2936,
  Minneapolis, Minnesota, June 2019. Association for Computational Linguistics.

\bibitem[Cobbe et~al.(2021)Cobbe, Kosaraju, Bavarian, Hilton, Nakano, Hesse,
  and Schulman]{cobbe2021training}
Cobbe, K., Kosaraju, V., Bavarian, M., Hilton, J., Nakano, R., Hesse, C., and
  Schulman, J.
\newblock Training verifiers to solve math word problems.
\newblock \emph{ArXiv preprint}, abs/2110.14168, 2021.

\bibitem[Dao et~al.(2022{\natexlab{a}})Dao, Fu, Ermon, Rudra, and
  R{\'e}]{dao2022flashattention}
Dao, T., Fu, D.~Y., Ermon, S., Rudra, A., and R{\'e}, C.
\newblock Flash{A}ttention: Fast and memory-efficient exact attention with
  {IO}-awareness.
\newblock In \emph{Advances in Neural Information Processing Systems},
  2022{\natexlab{a}}.

\bibitem[Dao et~al.(2022{\natexlab{b}})Dao, Fu, Saab, Thomas, Rudra, and
  R{\'e}]{dao2022hungry}
Dao, T., Fu, D.~Y., Saab, K.~K., Thomas, A.~W., Rudra, A., and R{\'e}, C.
\newblock Hungry hungry hippos: Towards language modeling with state space
  models.
\newblock \emph{ArXiv preprint}, abs/2212.14052, 2022{\natexlab{b}}.

\bibitem[Dong et~al.(2023)Dong, Li, Dai, Zheng, Wu, Chang, Sun, Xu, Li, and
  Sui]{dong_survey}
Dong, Q., Li, L., Dai, D., Zheng, C., Wu, Z., Chang, B., Sun, X., Xu, J., Li,
  L., and Sui, Z.
\newblock A survey for in-context learning.
\newblock \emph{ArXiv preprint}, abs/2301.00234, 2023.

\bibitem[Gao et~al.(2020)Gao, Biderman, Black, Golding, Hoppe, Foster, Phang,
  He, Thite, Nabeshima, Presser, and Leahy]{Gao2020ThePA}
Gao, L., Biderman, S.~R., Black, S., Golding, L., Hoppe, T., Foster, C., Phang,
  J., He, H., Thite, A., Nabeshima, N., Presser, S., and Leahy, C.
\newblock The pile: An 800gb dataset of diverse text for language modeling.
\newblock \emph{ArXiv}, abs/2101.00027, 2020.

\bibitem[Gao et~al.(2021)Gao, Fisch, and Chen]{gao2021making}
Gao, T., Fisch, A., and Chen, D.
\newblock Making pre-trained language models better few-shot learners.
\newblock In \emph{Proceedings of the 59th Annual Meeting of the Association
  for Computational Linguistics and the 11th International Joint Conference on
  Natural Language Processing (Volume 1: Long Papers)}, pp.\  3816--3830,
  Online, 2021. Association for Computational Linguistics.

\bibitem[Geva et~al.(2021)Geva, Khashabi, Segal, Khot, Roth, and
  Berant]{geva2021did}
Geva, M., Khashabi, D., Segal, E., Khot, T., Roth, D., and Berant, J.
\newblock Did aristotle use a laptop? a question answering benchmark with
  implicit reasoning strategies.
\newblock \emph{Transactions of the Association for Computational Linguistics},
  9:\penalty0 346--361, 2021.

\bibitem[Gu et~al.(2022)Gu, Goel, and Re]{gu2022efficiently}
Gu, A., Goel, K., and Re, C.
\newblock Efficiently modeling long sequences with structured state spaces.
\newblock In \emph{International Conference on Learning Representations}, 2022.

\bibitem[Gupta et~al.(2022)Gupta, Gu, and Berant]{gupta2022diagonal}
Gupta, A., Gu, A., and Berant, J.
\newblock Diagonal state spaces are as effective as structured state spaces.
\newblock In Oh, A.~H., Agarwal, A., Belgrave, D., and Cho, K. (eds.),
  \emph{Advances in Neural Information Processing Systems}, 2022.

\bibitem[Ho et~al.(2019)Ho, Kalchbrenner, Weissenborn, and
  Salimans]{ho2019axial}
Ho, J., Kalchbrenner, N., Weissenborn, D., and Salimans, T.
\newblock Axial attention in multidimensional transformers.
\newblock \emph{ArXiv preprint}, abs/1912.12180, 2019.

\bibitem[Hovy et~al.(2001)Hovy, Gerber, Hermjakob, Lin, and
  Ravichandran]{trec2001ho}
Hovy, E., Gerber, L., Hermjakob, U., Lin, C.-Y., and Ravichandran, D.
\newblock Toward semantics-based answer pinpointing.
\newblock In \emph{Proceedings of the First International Conference on Human
  Language Technology Research}, HLT '01, pp.\  1–7, USA, 2001. Association
  for Computational Linguistics.

\bibitem[Joshi et~al.(2017)Joshi, Choi, Weld, and
  Zettlemoyer]{joshi2017triviaqa}
Joshi, M., Choi, E., Weld, D., and Zettlemoyer, L.
\newblock {T}rivia{QA}: A large scale distantly supervised challenge dataset
  for reading comprehension.
\newblock In \emph{Proceedings of the 55th Annual Meeting of the Association
  for Computational Linguistics (Volume 1: Long Papers)}, pp.\  1601--1611,
  Vancouver, Canada, 2017. Association for Computational Linguistics.

\bibitem[Khashabi et~al.(2018)Khashabi, Chaturvedi, Roth, Upadhyay, and
  Roth]{khashabi-etal-2018-looking}
Khashabi, D., Chaturvedi, S., Roth, M., Upadhyay, S., and Roth, D.
\newblock Looking beyond the surface: A challenge set for reading comprehension
  over multiple sentences.
\newblock In \emph{Proceedings of the 2018 Conference of the North {A}merican
  Chapter of the Association for Computational Linguistics: Human Language
  Technologies, Volume 1 (Long Papers)}, pp.\  252--262, New Orleans,
  Louisiana, June 2018. Association for Computational Linguistics.

\bibitem[Lee et~al.(2019)Lee, Lee, Kim, Kosiorek, Choi, and Teh]{lee2019set}
Lee, J., Lee, Y., Kim, J., Kosiorek, A.~R., Choi, S., and Teh, Y.~W.
\newblock Set transformer: {A} framework for attention-based
  permutation-invariant neural networks.
\newblock In Chaudhuri, K. and Salakhutdinov, R. (eds.), \emph{Proceedings of
  the 36th International Conference on Machine Learning, {ICML} 2019, 9-15 June
  2019, Long Beach, California, {USA}}, volume~97 of \emph{Proceedings of
  Machine Learning Research}, pp.\  3744--3753. {PMLR}, 2019.

\bibitem[Levesque et~al.(2012)Levesque, Davis, and Morgenstern]{wsc2012hector}
Levesque, H.~J., Davis, E., and Morgenstern, L.
\newblock The winograd schema challenge.
\newblock In \emph{Proceedings of the Thirteenth International Conference on
  Principles of Knowledge Representation and Reasoning}, KR'12, pp.\
  552–561. AAAI Press, 2012.
\newblock ISBN 9781577355601.

\bibitem[Liu et~al.(2022)Liu, Shen, Zhang, Dolan, Carin, and
  Chen]{liu2022makes}
Liu, J., Shen, D., Zhang, Y., Dolan, B., Carin, L., and Chen, W.
\newblock What makes good in-context examples for {GPT}-3?
\newblock In \emph{Proceedings of Deep Learning Inside Out (DeeLIO 2022): The
  3rd Workshop on Knowledge Extraction and Integration for Deep Learning
  Architectures}, pp.\  100--114, Dublin, Ireland and Online, 2022. Association
  for Computational Linguistics.

\bibitem[Liu et~al.(2018)Liu, Saleh, Pot, Goodrich, Sepassi, Kaiser, and
  Shazeer]{liu2018generating}
Liu, P.~J., Saleh, M., Pot, E., Goodrich, B., Sepassi, R., Kaiser, L., and
  Shazeer, N.
\newblock Generating wikipedia by summarizing long sequences.
\newblock In \emph{6th International Conference on Learning Representations,
  {ICLR} 2018, Vancouver, BC, Canada, April 30 - May 3, 2018, Conference Track
  Proceedings}. OpenReview.net, 2018.

\bibitem[Lu et~al.(2021)Lu, Yao, Zhang, Zhu, Xu, Gao, XU, Xiang, and
  Zhang]{lu2021soft}
Lu, J., Yao, J., Zhang, J., Zhu, X., Xu, H., Gao, W., XU, C., Xiang, T., and
  Zhang, L.
\newblock Soft: Softmax-free transformer with linear complexity.
\newblock In Ranzato, M., Beygelzimer, A., Dauphin, Y., Liang, P., and Vaughan,
  J.~W. (eds.), \emph{Advances in Neural Information Processing Systems},
  volume~34, pp.\  21297--21309. Curran Associates, Inc., 2021.

\bibitem[Luong et~al.(2015)Luong, Pham, and Manning]{luong2015effective}
Luong, T., Pham, H., and Manning, C.~D.
\newblock Effective approaches to attention-based neural machine translation.
\newblock In \emph{Proceedings of the 2015 Conference on Empirical Methods in
  Natural Language Processing}, pp.\  1412--1421, Lisbon, Portugal, 2015.
  Association for Computational Linguistics.

\bibitem[Min et~al.(2022)Min, Lewis, Zettlemoyer, and
  Hajishirzi]{min-etal-2022-metaicl}
Min, S., Lewis, M., Zettlemoyer, L., and Hajishirzi, H.
\newblock {M}eta{ICL}: Learning to learn in context.
\newblock In \emph{Proceedings of the 2022 Conference of the North American
  Chapter of the Association for Computational Linguistics: Human Language
  Technologies}, pp.\  2791--2809, Seattle, United States, 2022. Association
  for Computational Linguistics.

\bibitem[Ott et~al.(2019)Ott, Edunov, Baevski, Fan, Gross, Ng, Grangier, and
  Auli]{ott2019fairseq}
Ott, M., Edunov, S., Baevski, A., Fan, A., Gross, S., Ng, N., Grangier, D., and
  Auli, M.
\newblock fairseq: A fast, extensible toolkit for sequence modeling.
\newblock In \emph{Proceedings of the 2019 Conference of the North {A}merican
  Chapter of the Association for Computational Linguistics (Demonstrations)},
  pp.\  48--53, Minneapolis, Minnesota, 2019. Association for Computational
  Linguistics.

\bibitem[Peng et~al.(2021)Peng, Pappas, Yogatama, Schwartz, Smith, and
  Kong]{peng2021random}
Peng, H., Pappas, N., Yogatama, D., Schwartz, R., Smith, N.~A., and Kong, L.
\newblock Random feature attention.
\newblock In \emph{9th International Conference on Learning Representations,
  {ICLR} 2021, Virtual Event, Austria, May 3-7, 2021}. OpenReview.net, 2021.

\bibitem[Peng et~al.(2022)Peng, Kasai, Pappas, Yogatama, Wu, Kong, Schwartz,
  and Smith]{peng2022abc}
Peng, H., Kasai, J., Pappas, N., Yogatama, D., Wu, Z., Kong, L., Schwartz, R.,
  and Smith, N.~A.
\newblock {ABC}: Attention with bounded-memory control.
\newblock In \emph{Proceedings of the 60th Annual Meeting of the Association
  for Computational Linguistics (Volume 1: Long Papers)}, pp.\  7469--7483,
  Dublin, Ireland, 2022. Association for Computational Linguistics.

\bibitem[Pilehvar \& Camacho-Collados(2019)Pilehvar and
  Camacho-Collados]{pilehvar-camacho-collados-2019-wic}
Pilehvar, M.~T. and Camacho-Collados, J.
\newblock {W}i{C}: the word-in-context dataset for evaluating context-sensitive
  meaning representations.
\newblock In \emph{Proceedings of the 2019 Conference of the North {A}merican
  Chapter of the Association for Computational Linguistics: Human Language
  Technologies, Volume 1 (Long and Short Papers)}, pp.\  1267--1273,
  Minneapolis, Minnesota, June 2019. Association for Computational Linguistics.

\bibitem[Radford et~al.(2019)Radford, Wu, Child, Luan, Amodei, and
  Sutskever]{radford2019language}
Radford, A., Wu, J., Child, R., Luan, D., Amodei, D., and Sutskever, I.
\newblock Language models are unsupervised multitask learners.
\newblock 2019.

\bibitem[Rae et~al.(2019)Rae, Potapenko, Jayakumar, Hillier, and
  Lillicrap]{raecompressive2019}
Rae, J.~W., Potapenko, A., Jayakumar, S.~M., Hillier, C., and Lillicrap, T.~P.
\newblock Compressive transformers for long-range sequence modelling.
\newblock \emph{arXiv preprint}, 2019.

\bibitem[Rae et~al.(2020)Rae, Potapenko, Jayakumar, Hillier, and
  Lillicrap]{rae2020Compressive}
Rae, J.~W., Potapenko, A., Jayakumar, S.~M., Hillier, C., and Lillicrap, T.~P.
\newblock Compressive transformers for long-range sequence modelling.
\newblock In \emph{8th International Conference on Learning Representations,
  {ICLR} 2020, Addis Ababa, Ethiopia, April 26-30, 2020}. OpenReview.net, 2020.

\bibitem[Rae et~al.(2021)Rae, Borgeaud, Cai, Millican, Hoffmann, Song,
  Aslanides, Henderson, Ring, Young, et~al.]{rae2021scaling}
Rae, J.~W., Borgeaud, S., Cai, T., Millican, K., Hoffmann, J., Song, F.,
  Aslanides, J., Henderson, S., Ring, R., Young, S., et~al.
\newblock Scaling language models: Methods, analysis \& insights from training
  gopher.
\newblock \emph{ArXiv preprint}, abs/2112.11446, 2021.

\bibitem[Raffel et~al.(2020)Raffel, Shazeer, Roberts, Lee, Narang, Matena,
  Zhou, Li, Liu, et~al.]{raffel2020exploring}
Raffel, C., Shazeer, N., Roberts, A., Lee, K., Narang, S., Matena, M., Zhou,
  Y., Li, W., Liu, P.~J., et~al.
\newblock Exploring the limits of transfer learning with a unified text-to-text
  transformer.
\newblock \emph{J. Mach. Learn. Res.}, 21\penalty0 (140):\penalty0 1--67, 2020.

\bibitem[Roemmele et~al.(2011)Roemmele, Bejan, and Gordon]{roemmele2011choice}
Roemmele, M., Bejan, C.~A., and Gordon, A.~S.
\newblock Choice of plausible alternatives: An evaluation of commonsense causal
  reasoning.
\newblock In \emph{2011 AAAI Spring Symposium Series}, 2011.

\bibitem[Scao et~al.(2022)Scao, Fan, Akiki, Pavlick, Ili{\'c}, Hesslow,
  Castagn{\'e}, Luccioni, Yvon, Gall{\'e}, et~al.]{scao-2022-bloom}
Scao, T.~L., Fan, A., Akiki, C., Pavlick, E., Ili{\'c}, S., Hesslow, D.,
  Castagn{\'e}, R., Luccioni, A.~S., Yvon, F., Gall{\'e}, M., et~al.
\newblock Bloom: A 176b-parameter open-access multilingual language model.
\newblock \emph{ArXiv preprint}, abs/2211.05100, 2022.

\bibitem[Smith et~al.(2022)Smith, Patwary, Norick, LeGresley, Rajbhandari,
  Casper, Liu, Prabhumoye, Zerveas, Korthikanti, Zhang, Child, Aminabadi,
  Bernauer, Song, Shoeybi, He, Houston, Tiwary, and
  Catanzaro]{Smith2022UsingDA}
Smith, S., Patwary, M., Norick, B., LeGresley, P., Rajbhandari, S., Casper, J.,
  Liu, Z., Prabhumoye, S., Zerveas, G., Korthikanti, V.~A., Zhang, E., Child,
  R., Aminabadi, R.~Y., Bernauer, J., Song, X., Shoeybi, M., He, Y., Houston,
  M., Tiwary, S., and Catanzaro, B.
\newblock Using deepspeed and megatron to train megatron-turing nlg 530b, a
  large-scale generative language model.
\newblock \emph{ArXiv}, abs/2201.11990, 2022.

\bibitem[Snell et~al.(2022)Snell, Klein, and Zhong]{snell2022learning}
Snell, C., Klein, D., and Zhong, R.
\newblock Learning by distilling context.
\newblock \emph{ArXiv preprint}, abs/2209.15189, 2022.

\bibitem[Socher et~al.(2013)Socher, Perelygin, Wu, Chuang, Manning, Ng, and
  Potts]{socher-etal-2013-recursive}
Socher, R., Perelygin, A., Wu, J., Chuang, J., Manning, C.~D., Ng, A., and
  Potts, C.
\newblock Recursive deep models for semantic compositionality over a sentiment
  treebank.
\newblock In \emph{Proceedings of the 2013 Conference on Empirical Methods in
  Natural Language Processing}, pp.\  1631--1642, Seattle, Washington, USA,
  October 2013. Association for Computational Linguistics.

\bibitem[Tay et~al.(2020)Tay, Bahri, Yang, Metzler, and Juan]{tay2020sparse}
Tay, Y., Bahri, D., Yang, L., Metzler, D., and Juan, D.
\newblock Sparse sinkhorn attention.
\newblock In \emph{Proceedings of the 37th International Conference on Machine
  Learning, {ICML} 2020, 13-18 July 2020, Virtual Event}, volume 119 of
  \emph{Proceedings of Machine Learning Research}, pp.\  9438--9447. {PMLR},
  2020.

\bibitem[Tay et~al.(2022)Tay, Tran, Dehghani, Ni, Bahri, Mehta, Qin, Hui, Zhao,
  Gupta, et~al.]{tay2022transformer}
Tay, Y., Tran, V.~Q., Dehghani, M., Ni, J., Bahri, D., Mehta, H., Qin, Z., Hui,
  K., Zhao, Z., Gupta, J., et~al.
\newblock Transformer memory as a differentiable search index.
\newblock \emph{ArXiv preprint}, abs/2202.06991, 2022.

\bibitem[Vaswani et~al.(2017)Vaswani, Shazeer, Parmar, Uszkoreit, Jones, Gomez,
  Kaiser, and Polosukhin]{vaswani2017attention}
Vaswani, A., Shazeer, N., Parmar, N., Uszkoreit, J., Jones, L., Gomez, A.~N.,
  Kaiser, L., and Polosukhin, I.
\newblock Attention is all you need.
\newblock In Guyon, I., von Luxburg, U., Bengio, S., Wallach, H.~M., Fergus,
  R., Vishwanathan, S. V.~N., and Garnett, R. (eds.), \emph{Advances in Neural
  Information Processing Systems 30: Annual Conference on Neural Information
  Processing Systems 2017, December 4-9, 2017, Long Beach, CA, {USA}}, pp.\
  5998--6008, 2017.

\bibitem[Wang et~al.(2020)Wang, Li, Khabsa, Fang, and Ma]{wang2020linformer}
Wang, S., Li, B., Khabsa, M., Fang, H., and Ma, H.
\newblock Linformer: Self-attention with linear complexity.
\newblock \emph{ArXiv preprint}, abs/2006.04768, 2020.

\bibitem[Wei et~al.(2021)Wei, Bosma, Zhao, Guu, Yu, Lester, Du, Dai, and
  Le]{FLAN2022Wei}
Wei, J., Bosma, M., Zhao, V.~Y., Guu, K., Yu, A.~W., Lester, B., Du, N., Dai,
  A.~M., and Le, Q.~V.
\newblock Finetuned language models are zero-shot learners.
\newblock \emph{ArXiv preprint}, abs/2109.01652, 2021.

\bibitem[Williams et~al.(2018)Williams, Nangia, and
  Bowman]{williams-etal-2018-broad}
Williams, A., Nangia, N., and Bowman, S.
\newblock A broad-coverage challenge corpus for sentence understanding through
  inference.
\newblock In \emph{Proceedings of the 2018 Conference of the North {A}merican
  Chapter of the Association for Computational Linguistics: Human Language
  Technologies, Volume 1 (Long Papers)}, pp.\  1112--1122, New Orleans,
  Louisiana, June 2018. Association for Computational Linguistics.

\bibitem[Wu et~al.(2022)Wu, Wang, Ye, and Kong]{wu2022self}
Wu, Z., Wang, Y., Ye, J., and Kong, L.
\newblock Self-adaptive in-context learning.
\newblock \emph{ArXiv preprint}, abs/2212.10375, 2022.

\bibitem[Xiong et~al.(2021)Xiong, Zeng, Chakraborty, Tan, Fung, Li, and
  Singh]{xiong2021nystromformer}
Xiong, Y., Zeng, Z., Chakraborty, R., Tan, M., Fung, G., Li, Y., and Singh, V.
\newblock Nyströmformer: A nyström-based algorithm for approximating
  self-attention.
\newblock \emph{Proceedings of the AAAI Conference on Artificial Intelligence},
  35\penalty0 (16):\penalty0 14138--14148, 2021.

\bibitem[Yang et~al.(2019)Yang, Dai, Yang, Carbonell, Salakhutdinov, and
  Le]{yang2019xlnet}
Yang, Z., Dai, Z., Yang, Y., Carbonell, J., Salakhutdinov, R.~R., and Le, Q.~V.
\newblock Xlnet: Generalized autoregressive pretraining for language
  understanding.
\newblock \emph{Advances in neural information processing systems}, 32, 2019.

\bibitem[Zaheer et~al.(2020)Zaheer, Guruganesh, Dubey, Ainslie, Alberti,
  Onta{\~{n}}{\'{o}}n, Pham, Ravula, Wang, Yang, and Ahmed]{zaheer2020big}
Zaheer, M., Guruganesh, G., Dubey, K.~A., Ainslie, J., Alberti, C.,
  Onta{\~{n}}{\'{o}}n, S., Pham, P., Ravula, A., Wang, Q., Yang, L., and Ahmed,
  A.
\newblock Big bird: Transformers for longer sequences.
\newblock In Larochelle, H., Ranzato, M., Hadsell, R., Balcan, M., and Lin, H.
  (eds.), \emph{Advances in Neural Information Processing Systems 33: Annual
  Conference on Neural Information Processing Systems 2020, NeurIPS 2020,
  December 6-12, 2020, virtual}, 2020.

\bibitem[Zhang et~al.(2022{\natexlab{a}})Zhang, Jiang, Feng, Zheng, and
  Kong]{zhang2022cab}
Zhang, J., Jiang, S., Feng, J., Zheng, L., and Kong, L.
\newblock Cab: Comprehensive attention benchmarking on long sequence modeling.
\newblock \emph{ArXiv preprint}, abs/2210.07661, 2022{\natexlab{a}}.

\bibitem[Zhang et~al.(2022{\natexlab{b}})Zhang, Roller, Goyal, Artetxe, Chen,
  Chen, Dewan, Diab, Li, Lin, Mihaylov, Ott, Shleifer, Shuster, Simig, Koura,
  Sridhar, Wang, and Zettlemoyer]{zhang_opt_2022}
Zhang, S., Roller, S., Goyal, N., Artetxe, M., Chen, M., Chen, S., Dewan, C.,
  Diab, M., Li, X., Lin, X.~V., Mihaylov, T., Ott, M., Shleifer, S., Shuster,
  K., Simig, D., Koura, P.~S., Sridhar, A., Wang, T., and Zettlemoyer, L.
\newblock {OPT}: Open pre-trained transformer language models.
\newblock \emph{ArXiv preprint}, abs/2205.01068, 2022{\natexlab{b}}.

\bibitem[Zhang et~al.(2022{\natexlab{c}})Zhang, Roller, Goyal, Artetxe, Chen,
  Chen, Dewan, Diab, Li, Lin, et~al.]{zhang2022opt}
Zhang, S., Roller, S., Goyal, N., Artetxe, M., Chen, M., Chen, S., Dewan, C.,
  Diab, M., Li, X., Lin, X.~V., et~al.
\newblock Opt: Open pre-trained transformer language models.
\newblock \emph{ArXiv preprint}, abs/2205.01068, 2022{\natexlab{c}}.

\bibitem[Zhang et~al.(2015)Zhang, Zhao, and LeCun]{zhang2015agnews}
Zhang, X., Zhao, J., and LeCun, Y.
\newblock Character-level convolutional networks for text classification.
\newblock In Cortes, C., Lawrence, N., Lee, D., Sugiyama, M., and Garnett, R.
  (eds.), \emph{Advances in Neural Information Processing Systems}, volume~28.
  Curran Associates, Inc., 2015.

\bibitem[Zheng et~al.(2022)Zheng, Wang, and Kong]{zheng2022linear}
Zheng, L., Wang, C., and Kong, L.
\newblock Linear complexity randomized self-attention mechanism.
\newblock In Chaudhuri, K., Jegelka, S., Song, L., Szepesvari, C., Niu, G., and
  Sabato, S. (eds.), \emph{Proceedings of the 39th International Conference on
  Machine Learning}, volume 162 of \emph{Proceedings of Machine Learning
  Research}, pp.\  27011--27041. PMLR, 2022.

\bibitem[Zheng et~al.(2023)Zheng, Yuan, Wang, and Kong]{zheng2023efficient}
Zheng, L., Yuan, J., Wang, C., and Kong, L.
\newblock Efficient attention via control variates.
\newblock In \emph{International Conference on Learning Representations}, 2023.
\newblock URL \url{https://openreview.net/forum?id=G-uNfHKrj46}.

\end{thebibliography}
\bibliographystyle{icml2023}

%%%%%%%%%%%%%%%%%%%%%%%%%%%%%%%%%%%%%%%%%%%%%%%%%%%%%%%%%%%%%%%%%%%%%%%%%%%%%%%
%%%%%%%%%%%%%%%%%%%%%%%%%%%%%%%%%%%%%%%%%%%%%%%%%%%%%%%%%%%%%%%%%%%%%%%%%%%%%%%
% APPENDIX
%%%%%%%%%%%%%%%%%%%%%%%%%%%%%%%%%%%%%%%%%%%%%%%%%%%%%%%%%%%%%%%%%%%%%%%%%%%%%%%
%%%%%%%%%%%%%%%%%%%%%%%%%%%%%%%%%%%%%%%%%%%%%%%%%%%%%%%%%%%%%%%%%%%%%%%%%%%%%%%
\newpage
\appendix
\onecolumn
\section{Pre-training Details}
\subsection{Data Processing}
\label{appendix:data}
We build the pre-training corpus based on the Pile~\citep{Gao2020ThePA}, and the pipeline includes filtering, deduplicating, and blending.
\paragraph{Filtering}
Many of our content filtering strategies were inspired by the data preparation pipeline of BLOOM~\citep{scao-2022-bloom} model.\footnote{\url{https://github.com/bigscience-workshop/data-preparation}}
We filtered raw data from the Pile, including catalog and content filtering.

For the catalog filtering, the pre-training corpus contains a subset of the Pile, including BookCorpus2, Books3, DM Mathematics, Project Gutenberg, HackerNews, OpenSubtitles, OpenWebText2, Pile-CC, USPTO, and Wikipedia.
We exclude the other subsets of the Pile.
On the one hand, based on this project's scope, we aim to demonstrate our model on the general natural language tasks, and the other domain-specific subsets of the Pile are unsuitable for this purpose.
On the other hand, these subsets are relatively noisy, which increases the difficulty and instabilities of the pre-training process, according to the tendency to cause spikes in gradient norms~\citep{zhang2022cab}.

For the content filtering, we first modified the raw data by standardizing the whitespace and removing the non-ASCII characters.
Then we filtered the text documents on (1) the flagged harmful words, (2) the stop word ratio, (3) the word/character repetition ratio, and (4) the specific character ratio.

\paragraph{Deduplicating}
We opted to take the fuzzy deduplication inspired by previous works~\citep{zhang_opt_2022, Smith2022UsingDA}.
In our implementation, we calculated the mini-hashes and performed LSH using \texttt{datasketch}\footnote{\url{https://github.com/ekzhu/datasketch}}, computed the connected components using \texttt{scipy}\footnote{\url{https://github.com/scipy/scipy}}, cached the hash fingerprint using \texttt{Redis}\footnote{\url{https://github.com/redis/redis}}.
We first whitespace-tokenized the documents into words and vectorized the documents with the 1-gram language model.
Then we calculated the mini-hashes of the document vectors to obtain the document fingerprints with 100-bit hash length.
We perform Locality Sensitive Hashing (LSH) through all the document fingerprints to find the neighborhoods of each document with a Jaccard similarity larger than 0.95.
After that, we constructed a sparse graph with each document as a node and connected the nodes with their neighborhoods.
In this way, we can find the sets of near-duplicated documents by computing the connected components of the graph.
Finally, we selected the high-quality documents from each set and removed the other documents in the order of predefined priority.

After the filtering and deduplication, we blended the filtered data into heterogeneous batches to obtain the final pre-training corpus. The details are shown in Table~\ref{appendix:table:ptcorpus}.

\begin{table*}[th]
\centering
\caption{Number of tokens per dataset in the final pre-training corpus}
\begin{tabular}{lc}
\toprule
\textbf{Datasets}          & \textbf{Tokens (billion)} \\
\midrule
BookCorpus2       & 1.6              \\
Gutenberg (PG-19) & 3.0              \\
Wikipedia (en)    & 12.1             \\
OpenWebText2      & 15.7             \\
Books3            & 26.0             \\
Pile-CC           & 52.2             \\
DM Mathematics    & 3.8              \\
HackerNews        & 1.1              \\
OpenSubtitles     & 1.6              \\
USPTO Backgrounds & 4.0              \\
\midrule
\textbf{Total}             & \textbf{121}             \\
\bottomrule
\end{tabular}

\label{appendix:table:ptcorpus}
\end{table*}

\subsection{Training Details}
\label{appendix:training}
We pre-trained \textbf{EvaLM} based on \texttt{metaseq}\footnote{\url{https://github.com/facebookresearch/metaseq/tree/main/metaseq}}, the pre-training hyperparameters are listed in~\tabref{appendix:table:pretrain-setting}.

\begin{table*}[th]
\centering
\caption{Hyperparameters used for pre-training}
\begin{tabular}{lcc}
\toprule
\textbf{Hypermeters}          
 & \textsc{EvaLM}-350M & \textsc{EvaLM}-1.3B \\
\midrule
Dropout      & \multicolumn{2}{c}{0.1}              \\
Weight Decay  & \multicolumn{2}{c}{0.1}                     \\
Clip Norm  & \multicolumn{2}{c}{1.0}                     \\
Clip Norm Type  & \multicolumn{2}{c}{L2}                     \\
LR Schedular  & \multicolumn{2}{c}{Polynomial decay}                     \\
Learning Rate    & \multicolumn{2}{c}{8e-5}  \\ 

Global Batch Size &64  &128 \\
DDP Backend     & DDP        & FSDP     \\
\bottomrule
\end{tabular}

\label{appendix:table:pretrain-setting}
\end{table*}
\section{Instruction Tuning Details}
We mainly follow settings in FLAN~\citep{FLAN2022Wei} to conduct ICL experiment. FLAN dataset consists of 12 dataset clusters including 9 NLU clusters and 3 NLG clusters. As we treat Agnews as a classification task, we only block out this dataset itself rather than the whole summarization cluster. 
The training hyperparameters are the same in the pre-training stage. We train all models for 5 epochs on selected FLAN datasets to get a fair comparison between IT, MSIT, and MSIT+ during the instruction tuning stage.
\label{appendix:IT}

% \begin{figure*}[ht]
%   \centering
%   \includegraphics[width=0.4\textwidth]{fig/ITshot1b3.pdf}
%   \caption{The accuracy on Trec dataset using top-k ICL, for 1.3B \textsc{EvaLM} models with different instruction tuning strategies.}
%   \label{fig:ITshot1b3}
% \end{figure*}

\section{In-Context Learning Results}
\subsection{In-context learning details}
We conduct in-context 
experiments with 0, 1, 3, 4, 8, 16, 32, 64, 80, 128, 192, 256, 372, 512, 640, 768, 896, 1024, 1280, 1536, 1792, 2000 shots considering the limited computing resources.

\label{appendix:icl details}

\begin{figure*}[htb]
\centering
\includegraphics[height=4.2cm]{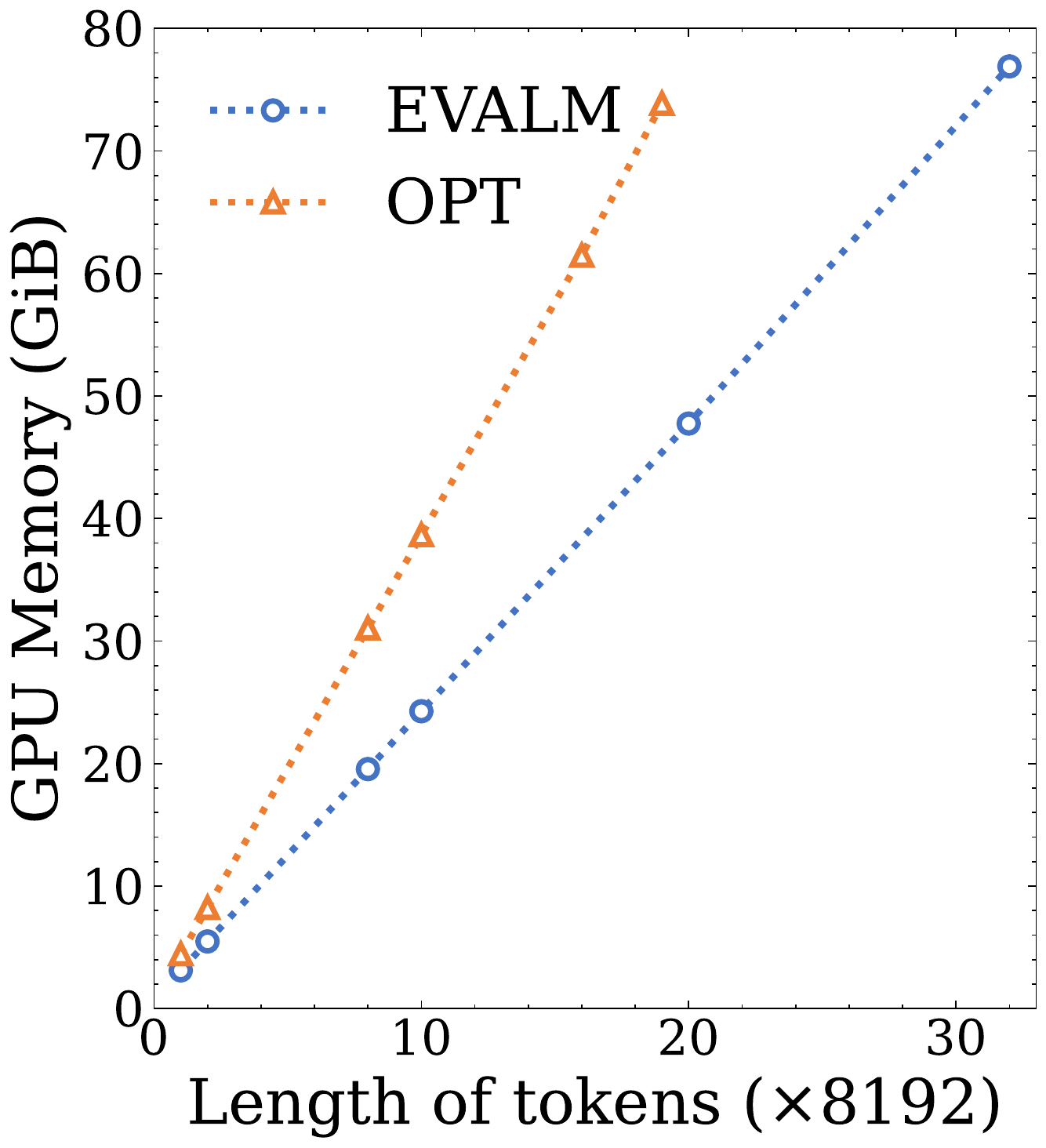}
\caption{Memory consumption for 350M models on A100 80G}
\label{fig:inf memo}
\end{figure*}

We compare the memory consumption for \textsc{EvaLM}-350M and OPT-350M on single NVIDIA 80G A100 in~\figref{fig:inf memo}.

We provide supplementary results for the many-shot setting in~\tabref{tab:icl-shots}, which is the shot number of demonstrations when the best score is achieved for each dataset respectively.

\begin{table*}[ht]
\small
\setlength\tabcolsep{2.3pt}
    \centering
    \caption{Supplementary results for many-shot setting: the shot number of demonstrations when the best score is achieved for each dataset respectively.}
    \begin{tabular}{l|cc|c|cc|cc|c|c|c}
    \toprule
    \multirow{2}{*}{\textbf{Models}} & \multicolumn{2}{c|}{Sentiment} & \multicolumn{1}{c|}{NLI}  & \multicolumn{2}{c|}{Miscellaneous}   & \multicolumn{2}{c|}{Reading}  & \multicolumn{1}{c|}{Topic} & \multicolumn{1}{c|}{Coreference}  &\multicolumn{1}{c}{CMS} \\  
    & SST-2  & SST-5   & MNLI    & Trec & WiC & MultiRC & BoolQ & AgNews  & WSC    & COPA \\
    \midrule
    OPT350M & 1 & 1 & 80  & 1 & 1 & 1 & 3 & 8 & 4 & 1 \\ 
         \textsc{EvaLM}350M & 1 & 4  & 372  & 372 & 128 & 8 & 4 & 64 & 16 & 1 \\ 
         \rowcolor{shadecolor}\quad w/ MSIT & 16 & 8  & 512  & 1280 & 128 & 16 & 3 & 80 & 256 & 64 \\
         \rowcolor{shadecolor}\quad w/ MSIT+ & 16 & 4 & 1280   & 372 & 128 & 64 & 3 & 80 & 8 & 128 \\
         OPT1.3B & 8 & 16 & 80  & 16 & 1 & 1 & 4 & 8 & 4 & 1 \\ 
         \textsc{EvaLM}1.3B & 192 & 16  & 256  & 256 & 192 & 8 & 128 & 64 & 372 & 16 \\
         \rowcolor{shadecolor}\quad w/ MSIT & 192 & 16 & 1280  & 256 & 192 & 16 & 128 & 64 & 512 & 16 \\ 
    \bottomrule
    \end{tabular}

    \label{tab:icl-shots}
\end{table*}

\subsection{Prompt Template}
\label{appendix:prompt}
For the sake of reproduction, we list the prompt template and label mapping used in our experiments for different tasks. We refer to templates protocol used in GPT3 and other works~\citep{wu2022self}.

\begin{table*}[th]
    \centering
    \label{tb:prompt-template}
    \caption{Prompt template and label mapping in our experiment}
    \begin{tabular}{llll}
    \toprule
    \textbf{Dataset} & \textbf{Template} & \textbf{Labal Space}\\
    \midrule
    SST-2 &  \{\textit{Label}\} Movie Review: \{\textit{Sentence}\} & Negative / Positive\\
    \midrule
    SST-5 & \{\textit{Sentence}\} It is \{\textit{Label}\} & terrible / bad / okay / good / great \\
    \midrule
    MNLI & \{\textit{Premise}\}?\{\textit{Label}\}, \{\textit{Hypothesis}\} &  No / Maybe / Yes\\
    \midrule
    % RTE & \{\textit{Premise}\}?\{\textit{Label}\}, \{\textit{Hypothesis}\} &  Yes / No\\
    Trec & \{\textit{Sentence}\} It is about \{\textit{Label}\} & 
    \tabincell{l}{abbreviation / entity / description and abstract \\concept / human being / location / numeric value}\\
    \midrule
    WIC & \tabincell{l}{\{\textit{Sentence1}\}$\backslash$n \{\textit{Sentence2}\}$\backslash$n\\ question: Is the word \{\textit{Word}\} used in the \\ same way in the two sentences above?$\backslash$n\\ answer: \{\textit{Label}\}} & no / yes\\
    \midrule
    MultiRC & \tabincell{l}{Context: \{\textit{Paragraph}\}$\backslash$n$\backslash$n \{\textit{Questions}\}$\backslash$n \\ \{\textit{Label}\} answer: \{\textit{Answer}\}} & incorrect / correct\\
    \midrule
    BoolQ & \tabincell{l}{Context:\{\textit{Passage}\}$\backslash$n Question: \{\textit{Question}\}?$\backslash$n\\answer: \{\textit{Label}\}} & no / yes\\
    \midrule
    AgNews & \{\textit{Sentence}\} It is about \{\textit{Label}\} & world / sports / business / technology\\
    \midrule
    WSC & \tabincell{l}{\{\textit{Paragraph}\}$\backslash$n Question: In the passage above, \\what does the pronoun \{\textit{Span2}\} refer to?$\backslash$n \\Answer:\{\textit{Span1}\} This is a \{\textit{Label}\} answer.} &  false / true \\
    \midrule
    COPA & \tabincell{l}{Context: \{\textit{Premise}\}$\backslash$n\\Correct Answer: \{\textit{Choices}\}} & false / true\\
    \bottomrule
    \end{tabular}
\end{table*}

\end{document}